\documentclass[10pt,twocolumn,letterpaper]{article}

\usepackage[pagenumbers]{wacv} 

\newcommand*{\V}[1]{\mathbf{#1}}

\newcommand{\refofpaper}[1]{of the main paper}
\newcommand{\refinpaper}[1]{in the main paper}

\usepackage{amsmath,amsfonts,bm}

\def\eqref#1{equation~\ref{#1}}

\def\1{\bm{1}}

\def\vx{{\bm{x}}}

\def\mI{{\bm{I}}}

\def\vx{{\bm{x}}}

\def\mI{{\bm{I}}}

\DeclareMathAlphabet{\mathsfit}{\encodingdefault}{\sfdefault}{m}{sl}
\SetMathAlphabet{\mathsfit}{bold}{\encodingdefault}{\sfdefault}{bx}{n}

\usepackage{url}
\usepackage{algorithm}
\usepackage{booktabs}
\usepackage{algpseudocode}
\usepackage{graphicx}
\usepackage{nicefrac}
\usepackage{tabularx}
\usepackage{multicol, multirow}
\usepackage{makecell}
\usepackage{verbatim}

\newcolumntype{Y}{>{\centering\arraybackslash}X}

\definecolor{wacvblue}{rgb}{0.21,0.49,0.74}
\usepackage[pagebackref,breaklinks,colorlinks,allcolors=wacvblue]{hyperref}

\title{\textit{Unconditional Priors Matter!} \\ Improving Conditional Generation of Fine-Tuned Diffusion Models}

\author{
Prin Phunyaphibarn \quad 
Phillip Y. Lee \quad 
Jaihoon Kim \quad 
Minhyuk Sung \\
KAIST \\
{\tt\small \{prin10517, phillip0701, jh27kim, mhsung\}@kaist.ac.kr}
}

\begin{document}
\twocolumn[{%
\renewcommand\twocolumn[1][]{#1}%
\maketitle
\begin{center}
    \centering
    \captionsetup{type=figure}
    \vspace{-\baselineskip}
    \includegraphics[width=0.9\textwidth]{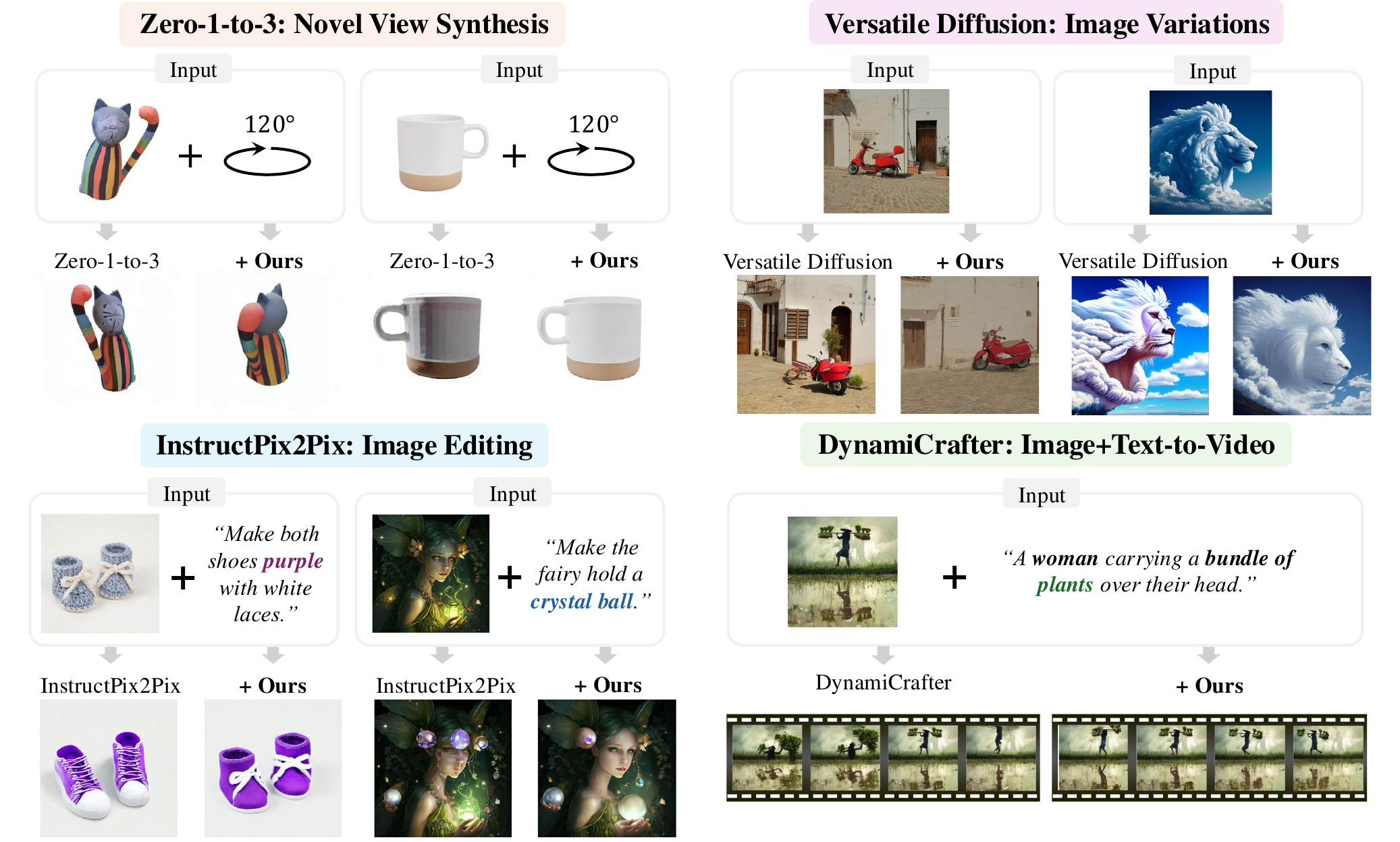}
    \vspace{-0.5\baselineskip}  
    \captionof{figure}{\textbf{Unconditional Priors Matter in CFG-Based Conditional Generation.} Fine-tuned conditional diffusion models often show drastic degradation in their unconditional priors, adversely affecting conditional generation when using techniques such as CFG~\cite{ho2021classifier}. We demonstrate that leveraging a diffusion model with a richer unconditional prior and combining its unconditional noise prediction with the conditional noise prediction from the fine-tuned model can lead to substantial improvements in conditional generation quality. This is demonstrated across diverse conditional diffusion models including Zero-1-to-3~\cite{liu2023zero}, Versatile Diffusion~\cite{xu2023versatile}, InstructPix2Pix~\cite{brooks2023instructpix2pix}, and DynamiCrafter~\cite{xing2025dynamicrafter}.} 
    \label{fig:teaser}
\end{center}
}]
\begin{abstract}
Classifier-Free Guidance (CFG) is a fundamental technique in training conditional diffusion models. The common practice for CFG-based training is to use a single network to learn both conditional and unconditional noise prediction, with a small dropout rate for conditioning. However, we observe that the joint learning of unconditional noise with limited bandwidth in training results in poor priors for the unconditional case. More importantly, these poor unconditional noise predictions become a serious reason for degrading the quality of conditional generation. Inspired by the fact that most CFG-based conditional models are trained by fine-tuning a base model with better unconditional generation, we first show that simply replacing the unconditional noise in CFG with that predicted by the base model can significantly improve conditional generation. 
Furthermore, we show that a diffusion model other than the one the fine-tuned model was trained on can be used for unconditional noise replacement. We experimentally verify our claim with a range of CFG-based conditional models for both image and video generation, including Zero-1-to-3, Versatile Diffusion, DiT, DynamiCrafter, and InstructPix2Pix.
\end{abstract}    
\vspace{-1.5\baselineskip}
\section{Introduction}
\vspace{-0.5\baselineskip}
In recent years, diffusion models~\cite{songscore, sohl2015deep, ho2020denoising} have shown great success in generation tasks, becoming the de facto standard generative model across many data modalities such as images~\cite{saharia2022photorealistic, peebles2023dit, rombach2022high, podell2023sdxl}, video~\cite{ho2022video, blattmann2023align, blattmann2023stable, zhang2024show}, and audio~\cite{huang2023make, copet2024simple, liu2024audioldm}. 
The success of diffusion models is not only due to their high-quality results and ease of training, but also the simplicity of adapting them into \emph{conditional} diffusion models. While previous generative models such as GANs~\cite{goodfellow2014generative} and VAEs~\cite{kingma2013auto} require separate training for each conditional generation task, making it costly to create various conditional generative models, diffusion models introduced a considerably more effective approach: training an unconditional model (or a conditional model with simple conditions, such as text) as a base and branching out into multiple conditional models.

At the core of the extendability of diffusion models in easily converting an unconditional (or less conditioned) base model into a conditional (or more conditioned) model is the Classifier-Free Guidance (CFG)~\cite{ho2021classifier} technique. CFG proposed to learn to predict both unconditional and conditional noises using a single neural network, without introducing another network, such as a classifier, as in the classifier-guidance~\cite{dhariwal2021diffusionbeatsgan} approach. CFG combines unconditional and conditional noise predictions to generate data conditioned on a given input. It has been widely adopted not only for training a conditional model from scratch but also for \emph{fine-tuning} a base model to incorporate other conditions, by adding encoders for the conditional input. Many successful conditional generative models have been fine-tuned using CFG from a base model. For example, Zero-1-to-3~\citep{liu2023zero} and Versatile Diffusion~\citep{xu2023versatile} use variants of Stable Diffusion~\cite{rombach2022high} (\texttt{SD}) as a base, with additional encoders to incorporate the input image as conditions, while InstructPix2Pix~\citep{brooks2023instructpix2pix} uses \texttt{SD1.5} as a base and incorporates text editing instructions and input reference images as conditions to perform instruction-based image editing.

Despite its successes and widespread usage, fine-tuning a conditional model from a base model using the CFG technique has limitations, most notably producing lower-quality results for unconditional generation. This is because both conditional and unconditional noise are learned by the same noise prediction network, thus sharing the limited capacity of the neural network. Typically, the bandwidth allocated for the unconditional noise is even more limited by setting a 5-20\% drop rate of the condition, an issue which is exacerbated when the training data is limited or the model is fine-tuned multiple times. More importantly, the low quality of unconditional noise also negatively affects the quality of conditional generation, since conditional generation is performed by combining both conditional and unconditional noise predictions in the CFG formulation.

The crucial oversight in this practice is that the base model already provides useful guidance for unconditional generation, and the quality of its generated outputs is generally much better than that of the fine-tuned model.
Hence, we demonstrate that in conditional generation using a fine-tuned diffusion model and CFG, simply replacing the unconditional noise of the fine-tuned diffusion model in CFG with that of the base model leads to significant improvements. This is a \emph{training-free} solution that requires no additional training or modifications to the neural networks. It also highlights that when fine-tuning a diffusion model with additional conditioning using CFG, the unconditional noise does not need to be learned jointly.

Surprisingly, we show that the unconditional noise does not need to come from the base unconditional model used for fine-tuning, but can also come from other pretrained diffusion models.
This further eliminates the need to jointly learn both unconditional and conditional noise using the same noise prediction network when a pretrained unconditional diffusion model is available.
\vspace{-0.5\baselineskip}
\section{Related Works}
\vspace{-0.25\baselineskip}

\paragraph{Guidance in Diffusion Models.}
Classifier-Free Guidance (CFG)~\cite{ho2021classifier} has become the de facto guidance technique for conditional generation with diffusion models, leading to notable improvements in both condition alignment and image quality. However, recent research has highlighted some of its limitations. \citet{kynkaanniemi2024applying} have shown that the specific timesteps at which CFG is applied significantly impact image diversity, and proposed to restrict CFG to certain intervals. 

Another line of work~\cite{hong2023improving, ahn2024self} addresses the limited applicability of CFG for \emph{text-based} conditions when using off-the-shelf diffusion models like Stable Diffusion~\cite{rombach2022high}. These approaches introduce a guidance technique that extends to a broader range of generation tasks, including unconditional generation, inverse problems, and conditional generation with \emph{non-text} conditions (\eg, depth maps~\cite{zhang2023adding}). However, most works have not explored how the dynamics of CFG shift when a diffusion model is \emph{fine-tuned} for a specific task~\cite{liu2023zero, xu2023versatile, brooks2023instructpix2pix}. Motivated by transfer learning, \citet{zhongdomain} propose to improve the generation quality of fine-tuned DiTs~\cite{peebles2023dit} by using the unconditional noise from the original model. However, they consider only small class-conditional DiTs while our method applies to popular large-scale diffusion models across different architectures (even when the base and fine-tuned models have different network architectures) and in cases where the base and fine-tuned model conditions have different modalities. Based directly on empirical observations of unconditional noise degradation in large-scale diffusion models that occurs during fine-tuning, we propose a solution by combining noise predictions from various diffusion models.

\citet{karras2024guiding} proposed Autoguidance which uses the noise estimate from an \emph{under-trained} version of itself, instead of unconditional noise, to improve image quality.
In Autoguidance, both versions of the model are conditioned on the \emph{same} condition to isolate the image quality improvement from the improvement in condition-alignment. In contrast, our method combines models whose conditions have \emph{different} modalities, correcting the degradation that stems from the quality of the unconditional prior used in the CFG formulation. Furthermore, Autoguidance emphasizes the importance of designing the degradations to match the degradation of the conditional model. They show that certain degradations will degrade rather than improve the generation quality. One of our main contributions is showing that the fine-tuned unconditional degradation hurts rather than helps the quality of conditional generation. 

\vspace{-0.5\baselineskip}
\paragraph{Merging Diffusion Models.}
Aligned with the mixture-of-experts~\cite{cai2024survey} and model merging~\cite{yang2024model} literature on foundation models, there is growing research on methods for merging diffusion models to enable effective composition of multiple conditions. Diffusion Soup~\cite{biggs2024diffusion} directly merges weights of different diffusion models, Mix-of-Show~\cite{gu2024mix} combines the weights of LoRA adapters~\cite{hu2021lora}, and MaxFusion~\cite{nair2025maxfusion} merges intermediate model features. Notably, leveraging the iterative denoising process of diffusion models, merging their noise estimates has emerged as a simple yet powerful technique for composing conditions. By merging noise estimates from the \emph{same} diffusion model with different input conditions, it becomes possible to generate outputs that contain a combination of these conditions~\cite{du2023reduce, du2024compositional, zhang2023diffcollage, bar2023multidiffusion, geng2024visual, geng2025factorized}. Interestingly, multiple studies have shown that noise estimates from \emph{different} diffusion models~\cite{nair2023unite, gandikota2023erasing, chen2024images} can also be merged effectively. In this work, we extend this approach, demonstrating how merging noise estimates can enhance generation quality when applying CFG to fine-tuned models.
\section{Background}
\label{sec:background}
\vspace{-0.25\baselineskip}
\subsection{Diffusion Models} 
\label{subsec:diffusion-models}
Diffusion models~\cite{ho2020denoising,songdenoising,sohl2015deep} generate data by sampling from a given distribution (\eg,~Gaussian) and applying iterative denoising. In the forward process, random noise is applied to clean data $\V{x}_0$ following:

\begin{align}
    \V{x}_t = \sqrt{\bar{\alpha}_t} \V{x}_0 + \sqrt{1 - \bar{\alpha}_t} \epsilon
\end{align}
where $\epsilon \sim \mathcal{N} (\mathbf{0}, \mathbf{I})$ and $\bar{\alpha}_t \in [0, 1]$\footnote{With variance schedule $ \{ \beta_t \}_{t=1}^{T}$, $\alpha_t:=1- \beta_t$ and $\bar{\alpha}_t := \prod_{s=1}^{t} \alpha_s$.}~\cite{ho2020denoising}. In the reverse process, the noisy data $\V{x}_t$ is denoised by modeling the transition as a Gaussian distribution:
\begin{align}
    p_{\theta} (\V{x}_{t-1} | \V{x}_{t}) = \mathcal{N} (\V{x}_{t-1}; \mu_\theta (\V{x}_t, t), \sigma^{2}_t \mathbf{I})
\end{align}
where the variance $\sigma^2_t$ is predefined, and predicting the posterior mean $\mu_\theta (\V{x}_t, t)$ can be reparameterized as a noise prediction task using Tweedie's formula~\cite{efron2011tweedie} as below: 
\begin{align}
    \label{eq:tweedies_formula}
    \mu_\theta (\V{x}_t, t) &= \Tilde{\mu}_t\left(\V{x}_t, \frac{1}{\sqrt{\bar{\alpha}}_t} (\V{x}_t - \sqrt{1 - \bar{\alpha}_t} \epsilon_\theta (\V{x}_t)) \right) \\
    &=: \Tilde{\mu} \left( \V{x}_t, g (\V{x}_t, \epsilon_{\theta} (\V{x}_t)) \right)
\end{align}
where $\epsilon_\theta (\V{x}_t)$\footnote{We omit timestep $t$ in $\epsilon_\theta (\V{x}_t)$ and $g(\V{x}_t, \epsilon_\theta (\V{x}_t))$ for brevity.} is the noise prediction from a diffusion model, and $\Tilde{\mu}_t(\V{x}_t, \V{x}_0)$ is the forward process posterior mean. Eq.~\ref{eq:tweedies_formula} can be further interpreted as updating the posterior mean towards the \emph{prediction of a clean observation} from $\V{x}_t$ using Tweedie's formula. We define this clean observation predicted from Tweedie's formula as the function $g(\cdot, \cdot)$ and denote the predicted clean observation by $\V{x}_{0 | t}$.

\vspace{-0.25\baselineskip}
\paragraph{DDIM Sampling.}
DDIM~\cite{songdenoising} enables efficient sampling for diffusion models by modeling the \emph{non-Markovian} transition $q (\V{x}_{t-1} | \V{x}_t, \V{x}_0)$, conditioned on $\V{x}_0$. One denoising step of DDIM is presented as the following deterministic transition:
\begin{align}
    \label{eq:ddim_denoising_step}
    \V{x}_{t-1} = \sqrt{\bar{\alpha}_{t-1}} g(\V{x}_t, \epsilon_\theta (\V{x}_t)) + \sqrt{1 - \bar{\alpha}_{t-1}} \epsilon_{\theta} (\V{x}_t).
\end{align}

\subsection{Classifier-Free Guidance (CFG)~\cite{ho2021classifier}} 
\label{subsec:CFG}
\vspace{-0.25\baselineskip}

For a diffusion model to perform \emph{conditional} generation given a condition $c$, it needs to sample from the conditional distribution $p(\V{x} | c)$. One approach is to use a classifier to guide the sampling process toward the conditional distribution~\cite{dhariwal2021diffusionbeatsgan}; however, it comes at the cost of training a separate classifier. Alternatively, \citet{ho2021classifier} eliminated the need for a separate classifier by introducing Classifier-Free Guidance (CFG), a straightforward modification to the training and sampling process of diffusion models. In CFG training, the model learns to predict the noise in $\V{x}_t$ at timestep $t$ not only when a condition $c$ is given, but also when a \emph{null condition} $\emptyset$ is given. That is, the diffusion model performs both \emph{conditional} (\ie,~$\epsilon_{\theta} (\V{x}_t, c)$) and \emph{unconditional} (\ie,~$\epsilon_{\theta} (\V{x}_t, \emptyset)$) noise prediction. This is achieved by setting the condition $c$ to the null condition $\emptyset$ with a certain probability during training. With a model trained using this technique, CFG can be applied in the sampling process by replacing $\epsilon_{\theta} (\V{x}_t)$ in Eq.~\ref{eq:ddim_denoising_step} with:
\begin{align}
    \label{eq:cfg-score}
    \epsilon_\theta^{(\gamma)}(\V{x}_t, c) = \epsilon_\theta (\V{x}_t, \emptyset) + \gamma (\epsilon_\theta (\V{x}_t, c) - \epsilon_\theta (\V{x}_t, \emptyset)),
\end{align}
where $\gamma$ is a guidance scale. A detailed algorithm of CFG with DDIM sampling~\cite{songdenoising} is shown in Alg.~\ref{alg:cfg-epsilon}.

\vspace{-0.5\baselineskip}
\paragraph{Analysis of CFG.}
Based on the connection between diffusion models and score-based models~\cite{songscore,song2019generative}, $\epsilon_\theta(\V{x}_t, c)$ and $\epsilon_\theta(\V{x}_t, \emptyset)$ model the conditional score $\nabla_{\V{x}_t} \log p(\V{x}_t | c)$ and the unconditional score $\nabla_{\V{x}_t} \log p(\V{x}_t)$ (up to a scaling factor), respectively. 
We can interpret the CFG noise $\epsilon_{\theta}^{(\gamma)} (\V{x}_t, c)$ in Eq.~\ref{eq:cfg-score} as an approximation of the true score $\nabla_{\V{x}_t} \log p_{\gamma}(\V{x}_t | c)$ where $p_{\gamma}(\V{x}_t | c) := p(\V{x}_t) \left( \frac{p(\V{x}_t | c)}{p(\V{x}_t)} \right)^\gamma$ is the \emph{gamma-powered} distribution. We can rewrite $p_{\gamma}(\V{x}_t | c)$ as follows:
\vspace{-0.5\baselineskip}
\begin{align*}
    p_{\gamma}(\V{x}_t | c) 
    = p(\V{x}_t) \left( \frac{p(\V{x}_t,c)}{p(c)p(\V{x}_t)} \right)^\gamma
    \propto p(\V{x}_t) p(c|\V{x}_t)^\gamma.
\end{align*}
Thus, CFG guides the samples via the implicit classifier $p(c|\V{x}_t)^\gamma$. Using $\gamma > 1$ results in sharpening the mode corresponding to $c$ which leads to better condition-alignment~\cite{ho2021classifier} and image quality~\citep{karras2024guiding}.
\begin{algorithm}[t!]
\caption{DDIM Sampling with CFG}\label{alg:cfg-epsilon}
\begin{algorithmic}[1]
\State $\mathbf{x}_T \sim \mathcal{N}(\mathbf{0},\mI)$
\For{$t = T, \dots, 1$}
    \State $\epsilon_\theta^{(\gamma)}(\V{x}_t, c) = \epsilon_\theta (\V{x}_t, \emptyset) + \gamma (\epsilon_\theta (\V{x}_t, c) - \epsilon_\theta (\V{x}_t, \emptyset))$ \label{ln:eps-cfg-score}
    \State \label{ln:eps-conditional-tweedies} $\V{x}_{0 | t} = g (\V{x}_t, \epsilon_\theta^{(\gamma)} (\V{x}_t, c))$
    \State \label{ln:eps-renoise} $\V{x}_{t-1} = \sqrt{\bar{\alpha}_{t-1}} \V{x}_{0|t} + \sqrt{1 - \bar{\alpha}_{t-1}} \epsilon^{(\gamma)}_{\theta} (\V{x}_t,c)$
\EndFor \\
\Return{$\mathbf{x}_0$}
\end{algorithmic}
\end{algorithm}

\vspace{-0.25\baselineskip}
\section{Unconditional Priors Matter}
\vspace{-0.25\baselineskip}
In this section, we discuss the negative impact of degraded unconditional noise estimates in diffusion models \emph{fine-tuned} on a narrower task-specific data distribution (Sec.~\ref{subsec:poor_uncond}). We then present a simple yet effective approach to enhance the generation quality of these models by leveraging richer unconditional noise estimates from other pretrained diffusion models (Sec.~\ref{subsec:good_uncond_base}).

\subsection{Poor Unconditional Priors Affect Conditional Generation}
\label{subsec:poor_uncond}
\vspace{-0.25\baselineskip}

The CFG training technique, introduced in Sec.~\ref{sec:background}, is also commonly used for \emph{fine-tuning} an existing diffusion model to incorporate new types of input conditions~\cite{liu2023zero, shi2023zero123++, sdiv-justin, xu2023versatile, brooks2023instructpix2pix}. Consider a pretrained diffusion model, such as Stable Diffusion~\cite{rombach2022high}, referred to as the \emph{base model}, parameterized by $\psi$. This base model can be fine-tuned on task-specific datasets to incorporate certain types of input conditions, such as camera poses~\cite{liu2023zero} or a reference image~\cite{xu2023versatile}. We refer to the resulting model as the \emph{fine-tuned} model, parameterized by $\theta$.

\begin{figure*}
    \centering
    \scriptsize
    \setlength{\tabcolsep}{0em}
    \def\arraystretch{0.0}
    {
    \begin{tabularx}{\linewidth}{Y | Y | Y | Y}
        \toprule
        Stable Diffusion v1.4~\cite{rombach2022high} & Versatile Diffusion~\cite{xu2023versatile}& Zero 1-to-3~\cite{liu2023zero} & InstructPix2Pix~\cite{brooks2023instructpix2pix} \\
        \midrule
        \multicolumn{4}{c}{\includegraphics[width=\textwidth]{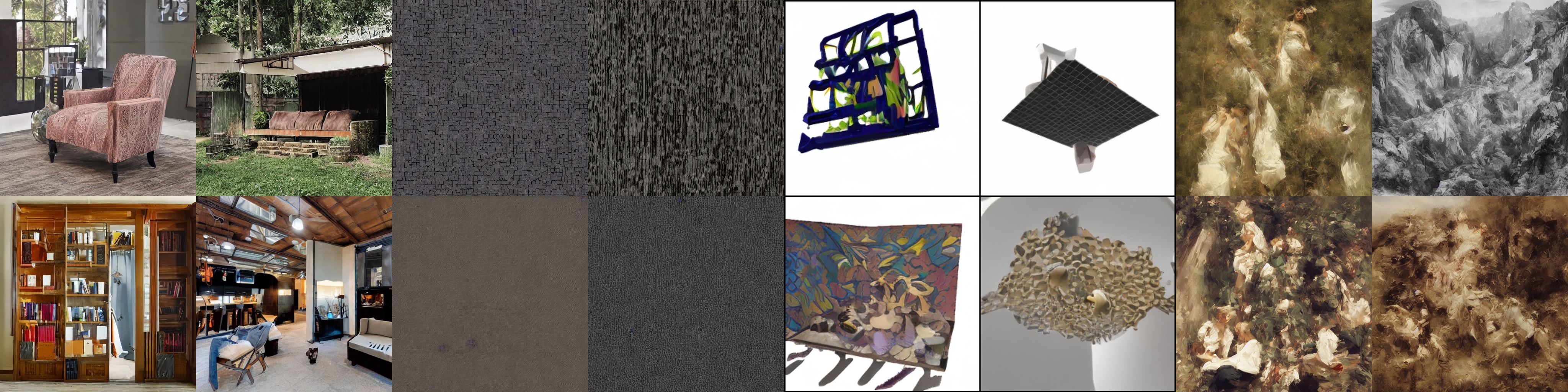}} \\
        \bottomrule
    \end{tabularx}
    }
    \vspace{-0.5\baselineskip}
    \caption{\textbf{Unconditional samples from different diffusion models.} Stable Diffusion~\cite{blattmann2023stable}, which often serves as the base model for fine-tuning conditional diffusion models, generates plausible images, whereas other fine-tuned diffusion models fail to sample realistic images.}
    \label{fig:unconditional-samples}
    \vspace{-1.0\baselineskip}
\end{figure*}

After fine-tuning the base model with CFG training, conditional generation with the fine-tuned model is performed by combining conditional and unconditional noise predictions following Eq.~\ref{eq:cfg-score}, yielding the CFG noise $\epsilon_\theta^{(\gamma)} (\V{x}_t, c)$. However, we observe a significant quality drop in \emph{unconditional} generation with the fine-tuned model compared to the base model. As shown in Fig.~\ref{fig:unconditional-samples}, the unconditional outputs from fine-tuned models~\cite{liu2023zero, xu2023versatile, brooks2023instructpix2pix} clearly lack detailed semantics and exhibit lower image quality. These results are expected, as (1) the unconditional distribution is inherently more complex than the conditional distribution, and (2) only a small fraction of the training data is utilized in each training iteration due to the low CFG dropping probability (typically 5-20\%).

Importantly, we observe that the quality of the \emph{conditional} generation in these fine-tuned models is negatively impacted by poor unconditional priors. This degradation can be understood through the CFG~\citep{ho2021classifier} formulation: CFG is designed to sample from the gamma-powered distribution $p_{\gamma}(\V{x}_t | c) \propto p(\V{x}_t) p(c | \V{x}_t)^\gamma$. Poor unconditional priors introduce approximation errors to $p(\V{x}_t)$, which in turn affects both $p(\vx_t)$ and $p(c|\vx_t) \propto \frac{p(\vx_t|c)}{p(\vx_t)}$. Although a key advantage of CFG is the joint modeling of the unconditional and conditional distributions, we find that under limited data or multiple fine-tuning, the fine-tuned model loses the rich unconditional prior of the base model, leading to quality degradation.

\subsection{Finding Richer Unconditional Priors}
\label{subsec:good_uncond_base}
\vspace{-0.25\baselineskip}
Can we improve the quality of conditional generation by incorporating better unconditional priors during sampling? Note that for fine-tuned diffusion models, we have access to a diffusion model with reliable unconditional priors: its base model. Therefore, we propose a simple yet effective fix, combining the \emph{unconditional} noise prediction from the base model with the \emph{conditional} noise prediction from the fine-tuned model. For this, the CFG noise in line~\ref{ln:eps-cfg-score} of Alg.~\ref{alg:cfg-epsilon} is modified as follows:
\begin{align}
    \label{eq:cfg-wr}
    {\color{purple}\epsilon_{\theta,\psi}^{(\gamma)}(\V{x}_t, c)} = {\color{teal} \epsilon_{\psi} (\V{x}_t, \emptyset)} + \gamma (\epsilon_{\theta}(\V{x}_t, c) -  {\color{teal} \epsilon_{\psi} (\V{x}_t, \emptyset)}),
\end{align}
where $\epsilon_\psi$ and $\epsilon_\theta$ denote the base model and fine-tuned model, respectively. Then DDIM sampling step becomes:
\vspace{-\baselineskip}
\begin{align*}
    \V{x}_{t-1} = \sqrt{\bar{\alpha}_{t-1}} g(\V{x}_t, {\color{purple}\epsilon_{\theta,\psi}^{(\gamma)}(\V{x}_t, c)}) + \sqrt{1 - \bar{\alpha}_{t-1}} {\color{purple}\epsilon_{\theta,\psi}^{(\gamma)}(\V{x}_t, c)}
\end{align*}

Surprisingly, this simple modification results in significant improvements in the output quality of conditional generation. We demonstrate this through both qualitative and quantitative evaluations in 
Sec.~\ref{sec:results}.

A natural next question that arises is: should the unconditional noise come from the base model or from another unconditional model? We find that the base model does not necessarily need to be the true base model from which the new conditional model was fine-tuned from, but can instead be another diffusion model with good unconditional priors. In our experiments, we show that even though some models have been fine-tuned on \texttt{SD1.x}, using unconditional predictions of \texttt{SD2.1} or \texttt{PixArt-$\alpha$} results in further improvements as shown in Sec.~\ref{sec:results}.

\vspace{-0.5\baselineskip}
\paragraph{Combining Diffusion Models.} Although the base model and the fine-tuned model have different model weights, their noise predictions can be combined as done in previous works~\cite{nair2023unite, gandikota2023erasing, chen2024images}. Based on the connection between energy-based models (EBMs) and diffusion models~\cite{liu2022compositional}, at each timestep, $t$, our method is equivalent to sampling from the time-annealed distribution $p_\psi(\V{x}_t)^{1-\gamma} p_\theta(\V{x}_t | c)^\gamma$ where $p_\psi(\V{x}_t)$ and $p_\theta(\V{x}_t)$ are the distributions modeled by the base and fine-tuned models, respectively. 

Notably, $p_\psi(\V{x}_t)$ can be modeled by any pretrained diffusion model which may have different weights or even different architecture from the fine-tuned model so long as $p_\psi(\V{x}_t)$ is a better approximation of the \emph{true} unconditional distribution than $p_\theta(\V{x}_t)$.
\section{Experiments}
\label{sec:results}
\vspace{-0.25\baselineskip}
We validate our method on five conditional diffusion models, each trained for distinct conditional generation tasks: Zero-1-to-3~\citep{liu2023zero}, Versatile Diffusion (VD)~\citep{xu2023versatile}, DiT~\cite{peebles2023dit}, DynamiCrafter~\citep{xing2025dynamicrafter}, and InstructPix2Pix~\citep{brooks2023instructpix2pix}.
For experiments on models fine-tuned from Stable Diffusion, we present the results using unconditional noise predictions from both the true base model of the fine-tuned networks and other diffusion models, Stable Diffusion 2.1 (\texttt{SD2.1})~\cite{rombach2022high} and \texttt{PixArt-$\alpha$}~\cite{chen2023pixart}. Notably, even when the fine-tuned model is  a \emph{UNet}, our method yields improvements when using \texttt{PixArt-$\alpha$}, which is a \emph{DiT}, in place of the base model. We refer readers to \textbf{supplementary} for details on the experimental setup for each application.

\subsection{Single-Condition CFG Formulation}
In this section, we provide experimental results of our method when applied to diffusion models that use single-condition CFG formulation. Models of this category samples using noise of the form provided in Eq.~\ref{eq:cfg-score}.
\paragraph{Zero-1-to-3~\citep{liu2023zero}}
\vspace{-0.25\baselineskip}

\begin{figure*}
    \centering
    \setlength{\tabcolsep}{0pt}
    \scriptsize
    \begin{tabularx}{\linewidth}{Y | Y | Y Y Y Y}
        \toprule
        \makecell{Input\\Image} & \makecell{Ground\\Truth} & \makecell{Zero-1-to-3\\\textbf{(Baseline)}} & \makecell{w/ \texttt{SD1.4}\\\textbf{(Ours)}} & \makecell{w/ \texttt{SD2.1}\\\textbf{(Ours)}} & \makecell{w/ \texttt{PixArt-$\alpha$}\\\textbf{(Ours)}} \\
        \midrule
        \multicolumn{6}{c}{\includegraphics[width=\linewidth]{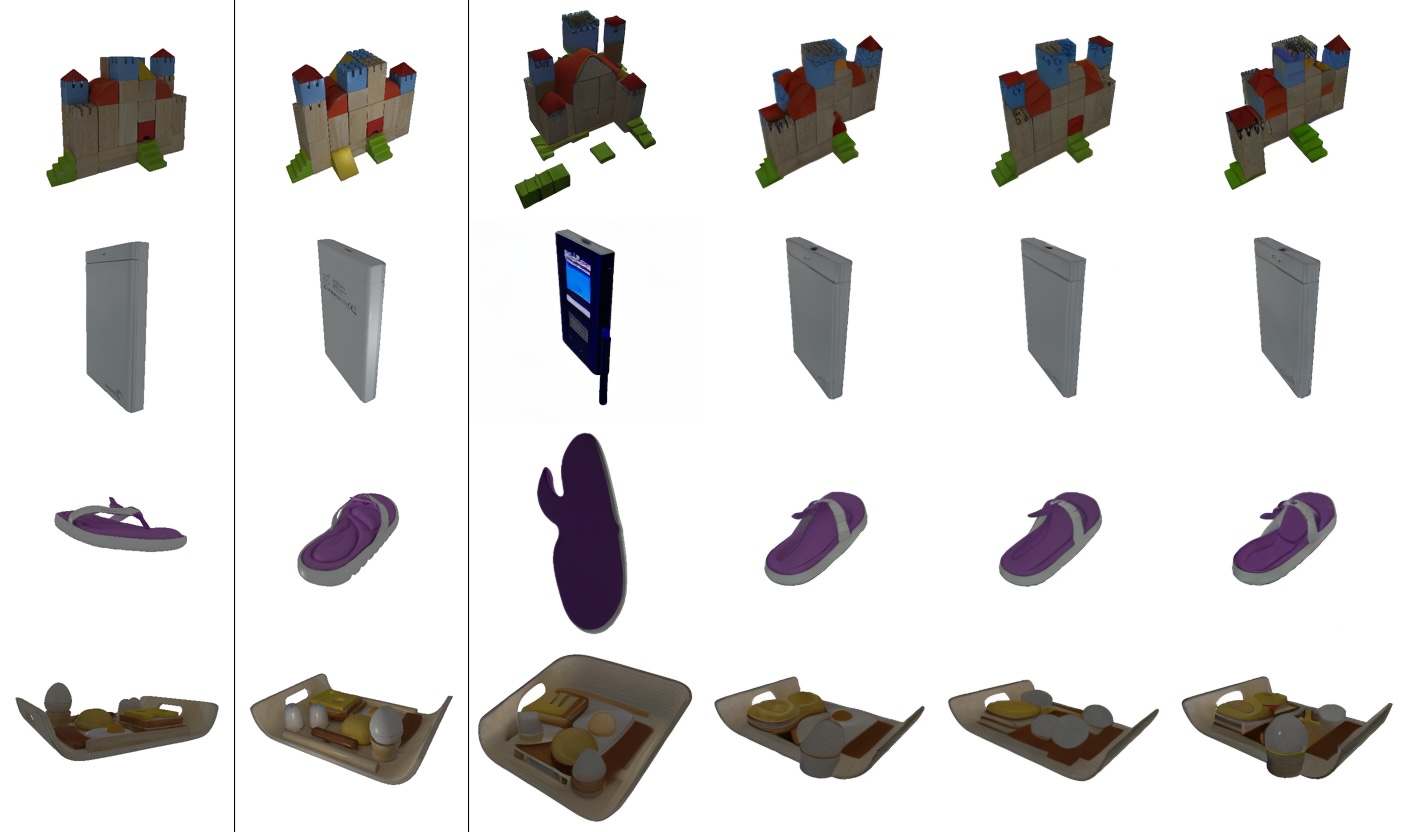}} \\
        \bottomrule
    \end{tabularx}
    \vspace{-\baselineskip}
    \caption{\textbf{Novel View Synthesis with Zero-1-to-3~\cite{liu2023zero}}. Outputs from Zero-1-to-3 often show inaccuracies in lighting or shape distortions during novel view synthesis. By incorporating unconditional noise predictions from Stable Diffusion~\cite{rombach2022high} or PixArt-$\alpha$~\cite{chen2023pixart}, our method achieves clear improvements in output quality.}
    \label{fig:zero123-main-qualitatives}
    \vspace{-\baselineskip}
\end{figure*}

Zero-1-to-3 is a conditional diffusion model for novel view synthesis, taking a reference image and relative camera poses as input. It is fine-tuned from Stable Diffusion Image Variations (\texttt{SD-IV})~\cite{sdiv-justin}, which itself is originally fine-tuned from \texttt{SD1.4}. 
Due to the multiple fine-tuning stages, we opted to use \texttt{SD1.4} as the base model.
We evaluate the samples on the Google Scanned Objects (GSO) dataset~\citep{gso} using LPIPS~\citep{zhang2018perceptual}, PSNR, and SSIM. As shown in Tab.~\ref{tab:zero123-results}, our method, which incorporates unconditional noise predictions from base models, achieves significant improvements across all three metrics, with the best performances observed when using \texttt{SD2.1} as the unconditional prior. Fig.~\ref{fig:zero123-main-qualitatives} shows that using improved unconditional noise from the base model enhances lighting quality (row 1), reduces color saturation (row 2) and shape distortions (rows 3 and 4).

\begin{table}
\begin{minipage}{\linewidth}
\centering
\scriptsize
\begin{tabularx}{\linewidth}{>{\centering\arraybackslash}m{0.3\textwidth} | Y | Y | Y}
\toprule
Method & LPIPS$\downarrow$ & PSNR$\uparrow$ & SSIM$\uparrow$ \\
\midrule
\scriptsize Zero-1-to-3~\cite{liu2023zero} & 0.182 & 16.647 & 0.824 \\
\textbf{Ours} w/ \texttt{SD1.4} & \underline{0.163} & \underline{17.514} & \underline{0.842} \\
\textbf{Ours} w/ \texttt{SD2.1} & \textbf{0.158} & \textbf{17.801} & \textbf{0.848} \\
\textbf{Ours} w/ \texttt{PixArt-$\alpha$} & 0.169 & 17.069 & 0.825 \\
\bottomrule
\end{tabularx}
\caption{\textbf{Novel View Synthesis with Zero-1-to-3~\citep{liu2023zero}.} Our method improves quality of novel view images (\textbf{bold} represents the best, and \underline{underline}, the second best method).}
\label{tab:zero123-results}
\vspace{-1.0\baselineskip}
\end{minipage}
\end{table}

\begin{table}
    \begin{minipage}{\linewidth}
    \centering
    \scriptsize
        \begin{tabularx}{\linewidth}{>{\centering\arraybackslash}m{0.3\textwidth} | >{\centering\arraybackslash}m{0.08\textwidth} | >{\centering\arraybackslash}m{0.14\textwidth} | >{\centering\arraybackslash}m{0.11\textwidth} | >{\centering\arraybackslash}m{0.13\textwidth}}
        \toprule
        Method & FID $\downarrow$ & $\text{FD}_{\text{DINOv2}}$ $\downarrow$ & CLIP-I $\uparrow$ & DINOv2 $\uparrow$ \\
        \midrule
        VD~\cite{xu2023versatile} & 8.38 & 167.65 & 0.93 & 0.91  \\
        \textbf{Ours} w/ \texttt{SD1.4} & \underline{6.68} & 156.77 & \textbf{0.94} & \textbf{0.92} \\
        \textbf{Ours} w/ \texttt{SD2.1} & 7.80 & \underline{151.48} & \textbf{0.94} & \textbf{0.92} \\
        \textbf{Ours} w/ \texttt{PixArt-$\alpha$} & \textbf{6.29} & \textbf{148.48} & \textbf{0.94} & \textbf{0.92} \\
        \bottomrule
    \end{tabularx}
    \vspace{-1.0\baselineskip}
    \caption{\textbf{Image Variations with Versatile Diffusion~\cite{xu2023versatile}.} Our sampling method achieves best performances across all metrics (\textbf{bold} represents the best, and \underline{underline}, the second best method).}
    \label{tab:vdiv-results}
    \vspace{-2.0\baselineskip}
    \end{minipage}
\end{table}

\vspace{-0.25\baselineskip}
\paragraph{Versatile Diffusion~\citep{xu2023versatile}}
\vspace{-0.25\baselineskip}
Versatile Diffusion (VD) is a multi-task diffusion model designed to handle text-to-image, image variations, and image-to-text tasks within a unified architecture. VD is progressively fine-tuned from \texttt{SD1.4} in three stages to handle additional image conditions on top of text condition. Due to the cascaded fine-tuning scheme, VD displays the worst image-unconditional generation quality as shown in Fig.~\ref{fig:unconditional-samples}. We focus on using VD for image variations to generate semantically similar images from a reference image.

We report FID~\cite{heusel2017gans} and $\text{FD}_\text{DINOv2}$~\cite{stein2024dgmeval} on COCO-Captions~\citep{lin2014microsoft} for image quality assessment and CLIP-I~\cite{hessel2021clipscore} and DINOv2~\cite{oquab2023dinov2} image similarity metrics to evaluate condition alignment.

As shown in Tab.~\ref{tab:vdiv-results}, using unconditional noise prediction from the base models yields better FID and $\text{FD}_\text{DINOv2}$ while retaining similar CLIP-I and DINOv2 image similarity, showing a performance improvement while maintaining condition alignment. 

\begin{figure*}
    \centering
    \scriptsize
    \setlength{\tabcolsep}{0pt}
    \begin{tabularx}{\textwidth}{Y | Y Y Y Y}
        \toprule
        {\makecell{Input Image}} & \makecell{VD\\\textbf{(Baseline)}} & \makecell{w/ \texttt{SD1.4}\\\textbf{(Ours)}} & \makecell{w/ \texttt{SD2.1}\\\textbf{(Ours)}} & \makecell{w/ \texttt{PixArt-$\alpha$}\\\textbf{(Ours)}} \\
        \midrule 
        \multicolumn{5}{c}{\includegraphics[width=\textwidth]{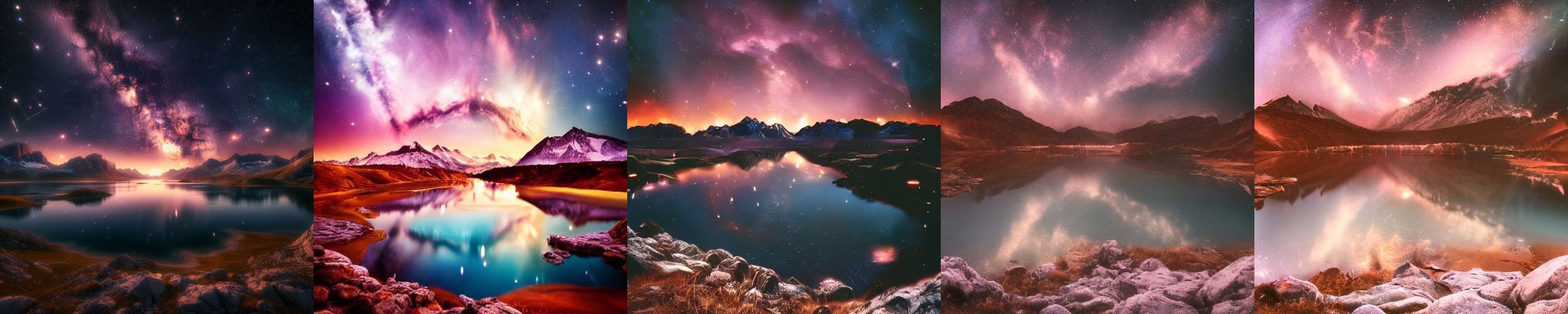}} \\
        \multicolumn{5}{c}{\includegraphics[width=\textwidth]{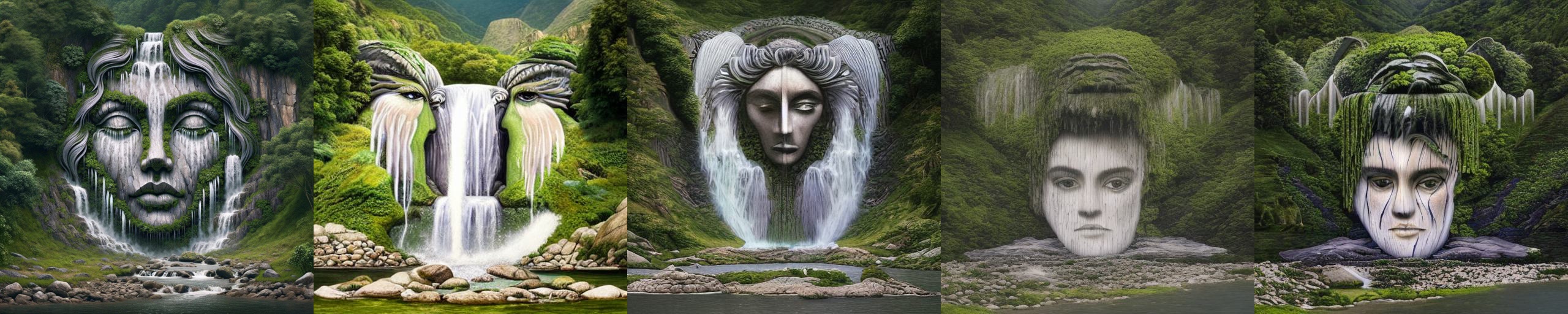}} \\
        \multicolumn{5}{c}{\includegraphics[width=\textwidth]{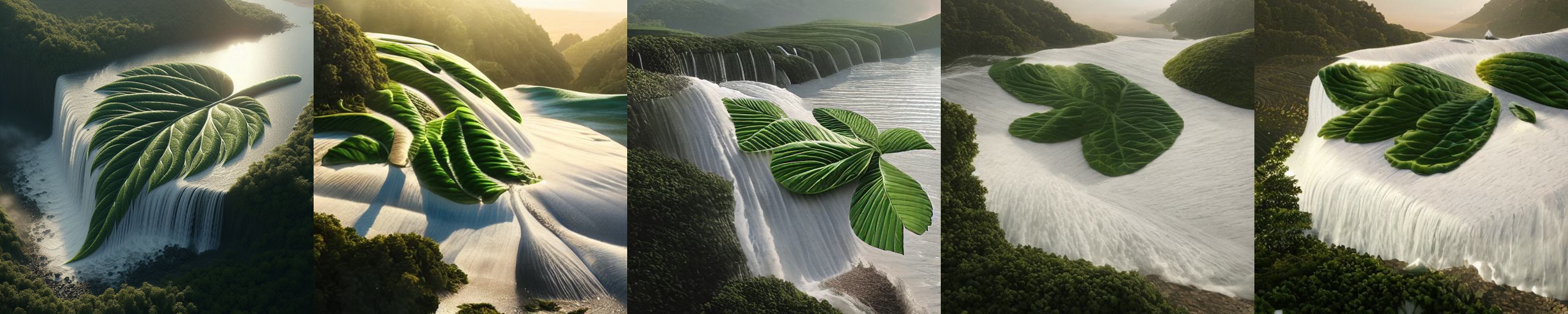}} \\
        \bottomrule
    \end{tabularx}
    \vspace{-\baselineskip}
    \caption{\textbf{Image Variations with Versatile Diffusion~\cite{xu2023versatile}.} Versatile Diffusion often suffers from style and detail degradation---excessive saturation (rows 1 and 3) or loss of key content (row 2). In contrast, our method, leveraging \texttt{SD1.4}, \texttt{SD2.1}, or \texttt{PixArt-$\alpha$} as unconditional priors, achieves noticeable improvements in performance.}
    \label{fig:vdiv-selected-qualitatives}
    \vspace{-1.5\baselineskip}
\end{figure*}

As shown in Fig.~\ref{fig:vdiv-selected-qualitatives}, VD often generates images with highly saturated colors (rows 1 and 2) and distorted objects (row 3) while our method corrects both. 

\vspace{-\baselineskip}
\paragraph{DiT~\cite{peebles2023dit}}
Although the experiments so far have been conducted using diffusion models with the UNet architecture, we show that our method holds for fine-tuned DiTs~\cite{peebles2023dit} as well. Since there are no publicly available fine-tuned DiT models, we fine-tune DiT-XL/2 on the standard downstream tasks SUN397~\cite{xiao2010sun}, Food101~\cite{bossard2014food}, and Caltech101~\cite{griffin2007caltech}. The FID for the different fine-tuning tasks are shown in Tab.~\ref{tab:dit-results}. Incorporating the unconditional noise from the base DiT-XL/2 results in improved FID. We also observe larger benefits when the fine-tuning dataset is large (Food101 and SUN397) corroborating our observation that the degradation in unconditional priors is amplified by the limited computational budget. While we observe no improvement for Caltech101, the dataset is over 10 times smaller than both SUN397 and Food101, thus the model is given sufficient time to fit the dataset despite the low CFG condition drop rate (10\%). In practice, large-scale fine-tuning like in Zero-1-to-3 and Versatile Diffusion often suffer from limited data and computation. In these practical settings, the degradation of the unconditional prior becomes highly detrimental as we have shown.

\begin{table}[h]
\begin{minipage}{\linewidth}
\centering
\scriptsize
\begin{tabularx}{\linewidth}{>{\centering\arraybackslash}m{0.3\textwidth} | Y Y Y}
\toprule
Method & SUN397 & Food101 & Caltech101 \\
\midrule
Fine-tuned DiT-XL/2 & 17.12 & 18.31 & \textbf{24.05} \\
\textbf{Ours} & \textbf{14.51} & \textbf{17.67} & 24.15 \\
\bottomrule
\end{tabularx}
\vspace{-0.5\baselineskip}
\caption{\textbf{Class-conditional generation with DiT\citep{peebles2023dit}.} FID-5k evaluated on three fine-tuning tasks (SUN397~\cite{xiao2010sun}, Food101~\cite{bossard2014food}, and Caltech101~\cite{griffin2007caltech}). Our method improves FID of the fine-tuned models (\textbf{bold} represents the best method).}
\label{tab:dit-results}
\vspace{-\baselineskip}
\end{minipage}
\end{table}

\subsection{Dual-Condition CFG Formulation}
In this section, we provide experimental results on diffusion models which use the dual-condition CFG formulation. Diffusion models in this category are conditioned on two conditions and sample using the modified CFG noise
\vspace{-0.5\baselineskip}
\begin{align}
    \epsilon_\theta(\V{x}_t, c_1, c_2) = &\;{\color{magenta} \epsilon_\theta(\V{x}_t, \emptyset, \emptyset)} \notag \\
    &+ \gamma_1 (\epsilon_\theta(\V{x}_t, c_1, \emptyset) - {\color{magenta}\epsilon_\theta(\V{x}_t, \emptyset, \emptyset)}) \notag \\
    &+ \gamma_2 (\epsilon_\theta(\V{x}_t, c_1, c_2) - \epsilon_\theta(\V{x}_t, c_1, \emptyset)). \label{eq:dual-cfg}
\end{align}
Notably, the dual-condition CFG formulation has two unconditional terms trained using CFG condition dropout: $\epsilon_\theta(\V{x}_t, \emptyset, \emptyset)$ and $\epsilon_\theta(\V{x}_t, c_1, \emptyset)$. We replace ${\color{magenta}\epsilon_\theta(\V{x}_t, \emptyset, \emptyset)}$ with the base model unconditional noise prediction $\epsilon_\psi(\V{x}_t, \emptyset)$. However, since the other unconditional term $\epsilon_\theta(\V{x}_t, c_1, \emptyset)$ is not replaced, we observe less improvements in this case than in the single-condition CFG formulation. This is to be expected as the quality degradation stems from training using a low dropout rate for the condition which is applied to both ${\color{magenta}\epsilon_\theta(\V{x}_t, \emptyset, \emptyset)}$ and $\epsilon_\theta(\V{x}_t, c_1, \emptyset)$, only one of which is replaced by the better base model unconditional prior.

\begin{figure}[h]
    \centering
    \setlength{\tabcolsep}{0pt}
    \scriptsize
    \vspace{-0.25\baselineskip}
    \begin{tabularx}{\linewidth}{>{\centering\arraybackslash}m{0.09\textwidth} | Y Y Y Y}
        \toprule
        Input & \multicolumn{4}{c}{\textbf{Generated Frames}} \\
        \midrule 
        \multicolumn{5}{c}{\includegraphics[width=\linewidth]{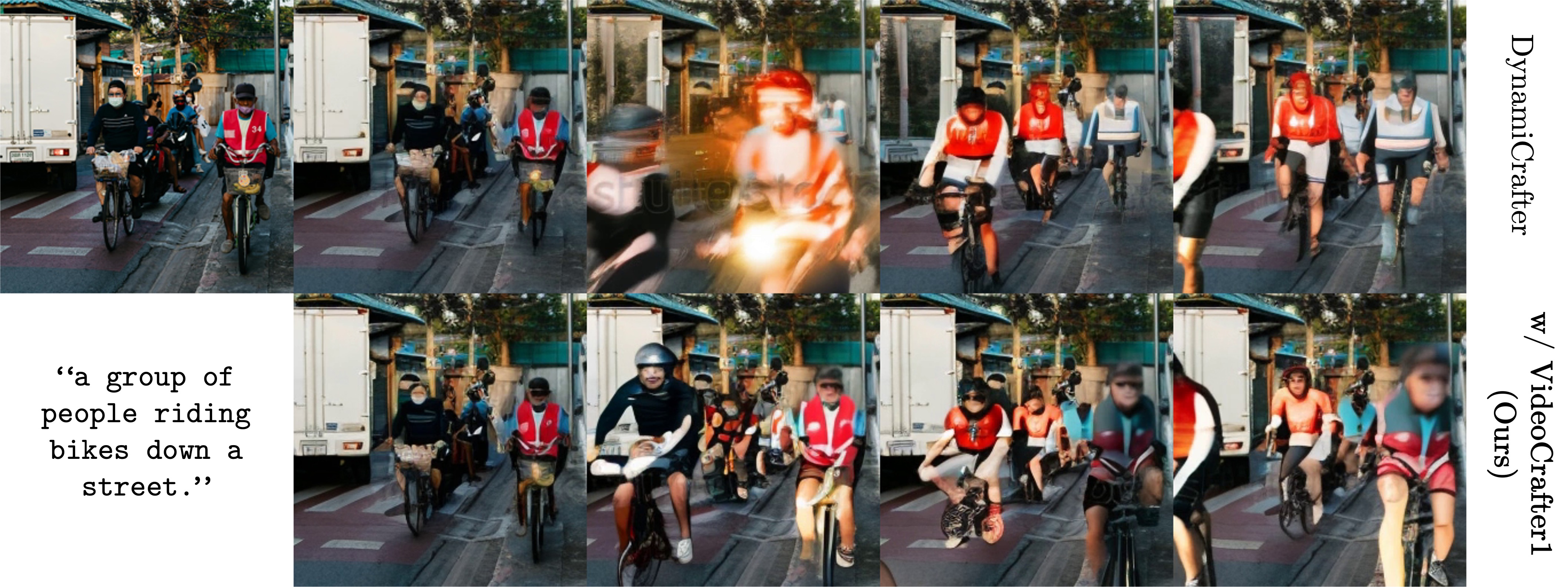}} \\
        \multicolumn{5}{c}{\includegraphics[width=\linewidth]{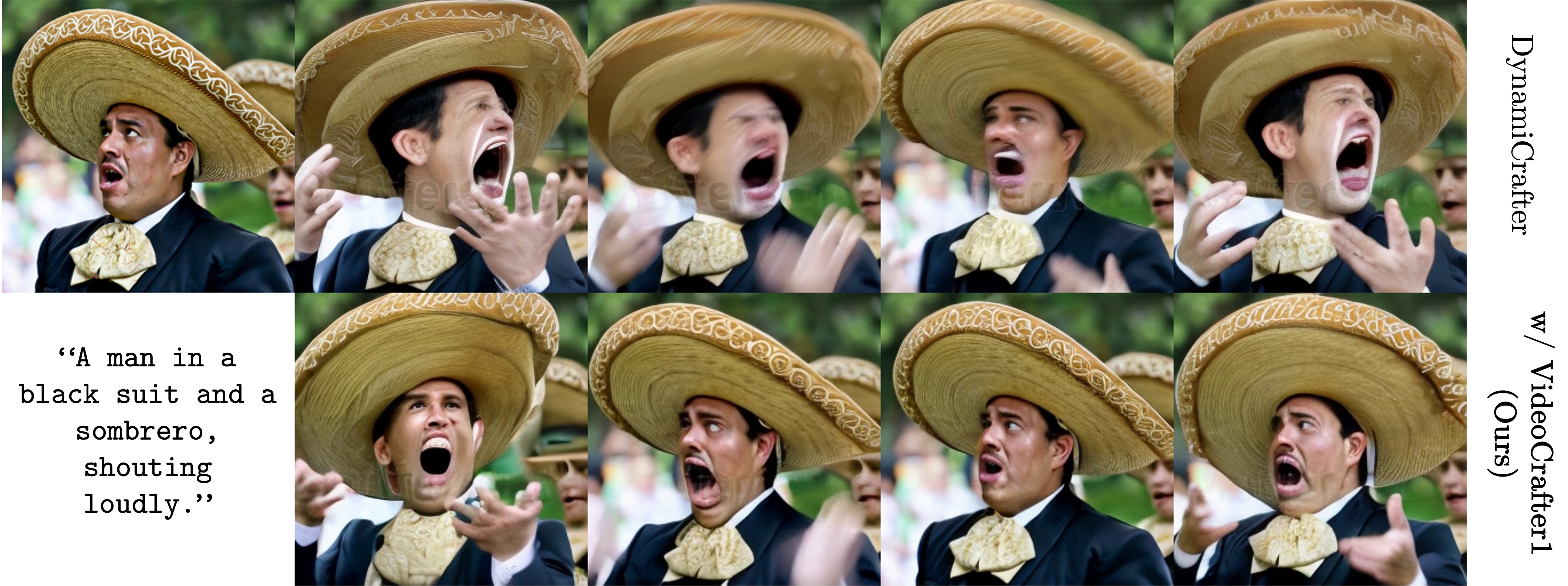}} \\
        \bottomrule
    \end{tabularx}
    \vspace{-\baselineskip}
    \caption{\textbf{Image-to-Video Generation with DynamiCrafter~\cite{xing2025dynamicrafter}.} Our method is more temporally consistent (lighting on the biker) and less distorted (the hand and face in the second video).}
    \label{fig:dynamicrafter-qualitatives}
    \vspace{-2\baselineskip}
\end{figure}

\vspace{-\baselineskip}
\paragraph{DynamiCrafter~\citep{xing2025dynamicrafter}}\label{sec:dynamicrafter}
We apply our method to DynamiCrafter~\cite{xing2025dynamicrafter}, a text-and-image-to-video diffusion model fine-tuned from the text-to-video diffusion model VideoCrafterT2V~\cite{chen2023videocrafter1}. DynamiCrafter incorporates an image condition $c_I$ as $c_1$ and a text condition $c_T$ as $c_2$. For our method, we replace the DynamiCrafter unconditional noise with the VideoCrafterT2V unconditional noise. We report quantitative results using VBenchI2V~\cite{huang2024vbench, huang2023vbenchgithub} which measures video quality and temporal consistency across multiple dimensions. For more details on the metrics, please refer to VBench~\cite{huang2024vbench}. Quantitative results are reported in Tab.~\ref{tab:dynamicrafter-results}. Our method outperforms the baseline in 7 out of 9 metrics, yielding more consistent videos with higher aesthetic quality. As shown in Fig.~\ref{fig:dynamicrafter-qualitatives}, our method is more temporally consistent (first video) and less distorted (second video). Video results are also included in the \textbf{supplementary}.

\begin{table*}[t]
    \begin{minipage}{\linewidth}
    \centering
    \scriptsize
        \begin{tabularx}{\linewidth}{>{\centering\arraybackslash}m{0.11\textwidth} | >{\centering\arraybackslash}m{0.08\textwidth} | >{\centering\arraybackslash}m{0.08\textwidth} | >{\centering\arraybackslash}m{0.08\textwidth} | >{\centering\arraybackslash}m{0.08\textwidth} | >
        {\centering\arraybackslash}m{0.06\textwidth} | >{\centering\arraybackslash}m{0.07\textwidth} | >{\centering\arraybackslash}m{0.07\textwidth} | >{\centering\arraybackslash}m{0.06\textwidth} | >{\centering\arraybackslash}m{0.07\textwidth}}
        \toprule
         Method & \makecell{Subject\\Consistency}  & \makecell{Background\\Consistency} & \makecell{Temporal\\Flickering} & \makecell{Motion\\Smoothness} & \makecell{Dynamic\\Degree} & \makecell{Aesthetic\\Quality} & \makecell{Imaging\\Quality} & \makecell{I2V\\Subject} & \makecell{I2V\\Background} \\
        \midrule
        DynamiCrafter~\cite{xing2025dynamicrafter}       & 90.80 & 96.73 & 95.21 & 96.67 & \textbf{59.59} & 57.06 & \textbf{64.10} & 92.77 & 94.56 \\
        \textbf{Ours}   & \textbf{91.49} & \textbf{97.03} & \textbf{95.34} & \textbf{96.86} & 57.32 & \textbf{57.51} & 63.15 & \textbf{93.49} & \textbf{94.72} \\
        \bottomrule
        \end{tabularx}
        \vspace{-0.5\baselineskip}
        \caption{\textbf{Video Generation with DynamiCrafter~\cite{xing2025dynamicrafter}.} All metrics are scored out of 100, higher indicates better performance.}
        \vspace{-\baselineskip}
    \label{tab:dynamicrafter-results}
    \end{minipage}
\end{table*}

\begin{figure}
    \centering
    \setlength{\tabcolsep}{0pt}
    \scriptsize
    \vspace{-0.25\baselineskip}
    \begin{tabularx}{\linewidth}{Y | Y Y Y Y}
        \toprule
        {\makecell{Input \\ Image}} & {\makecell{IP2P\\\textbf{(Baseline)}}} & {\makecell{w/ SD1.5\\\textbf{(Ours)}}} &
        {\makecell{w/ SD2.1\\\textbf{(Ours)}}} & {\makecell{w/ PixArt-$\alpha$\\\textbf{(Ours)}}}\\
        \midrule 
        \multicolumn{5}{c}{\scriptsize\texttt{``\input{figures/ip2p/piebench/10.txt}''}} \\
        \multicolumn{5}{c}{\includegraphics[width=\linewidth]{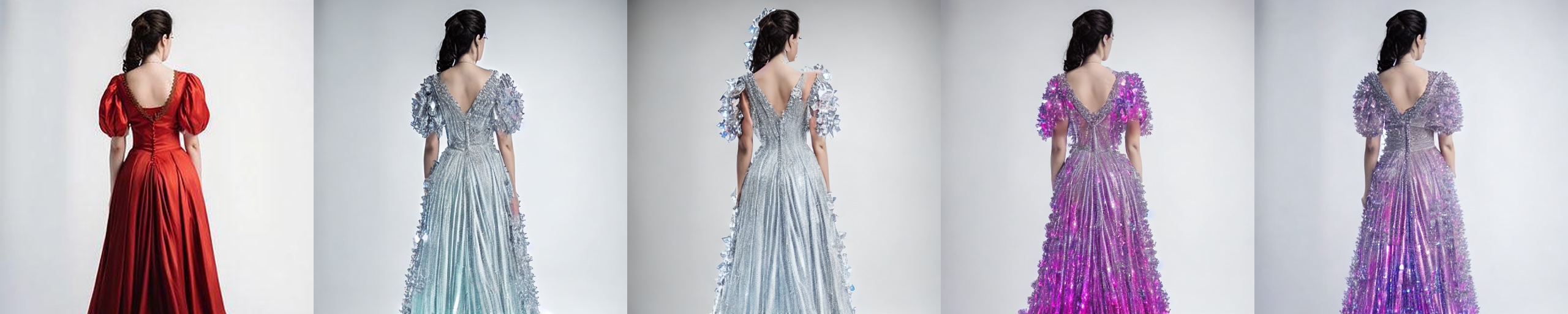}} \\
        \multicolumn{5}{c}{\scriptsize\texttt{``\input{figures/ip2p/piebench/08.txt}''}} \\
        \multicolumn{5}{c}{\includegraphics[width=\linewidth]{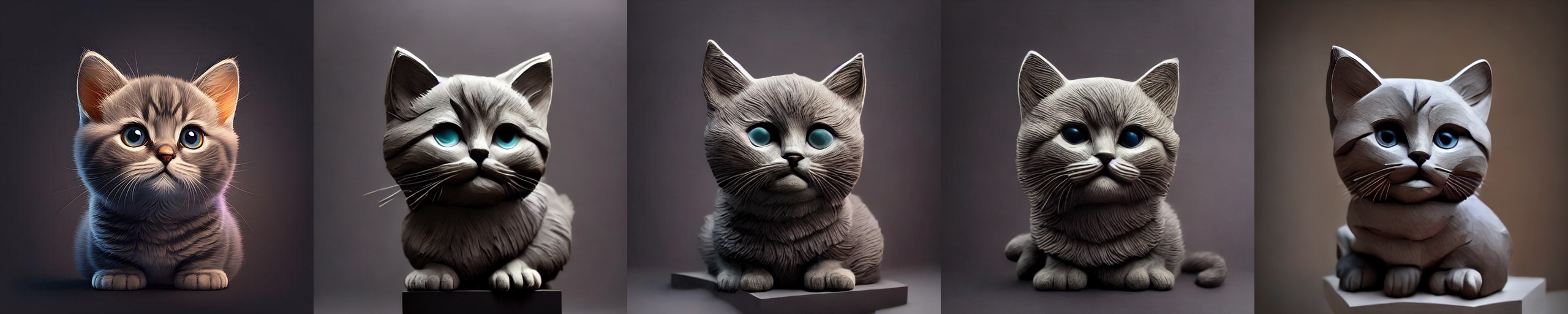}} \\
        \multicolumn{5}{c}{\scriptsize\texttt{``\input{figures/ip2p/piebench/01.txt}''}} \\
        \multicolumn{5}{c}{\includegraphics[width=\linewidth]{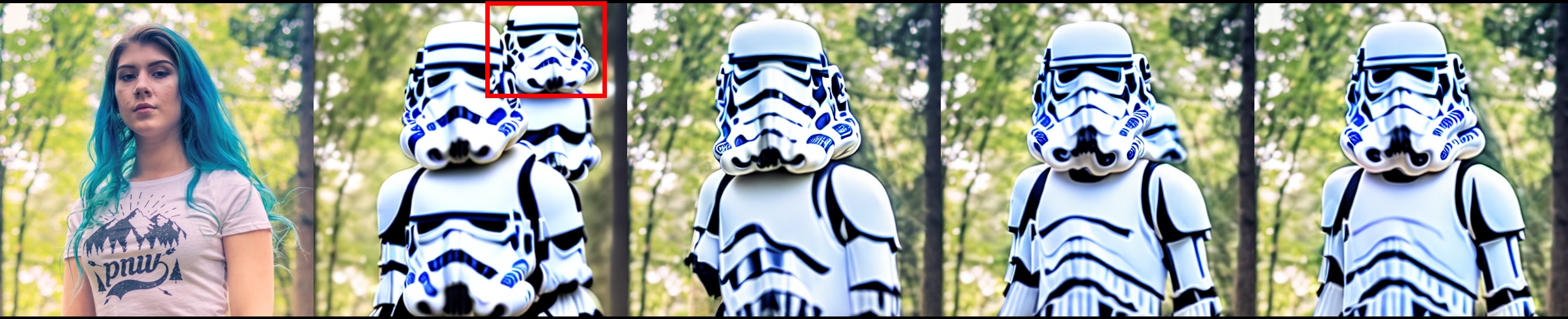}} \\
        \bottomrule
    \end{tabularx}
    \vspace{-\baselineskip}
    \caption{\textbf{Image Editing with InstructPix2Pix (IP2P)~\cite{brooks2023instructpix2pix}.} Applying our method improves alignment with the editing prompt while preserving the identity of the source image.}
    \label{fig:ip2p-selected-qualitatives}
    \vspace{-1.5\baselineskip}
\end{figure}

\vspace{-\baselineskip}
\paragraph{InstructPix2Pix~\citep{brooks2023instructpix2pix}}
InstructPix2Pix (IP2P) tackles instruction-based image editing by fine-tuning \texttt{SD1.5} to condition on both text (editing instruction, as $c_2)$ and image (as $c_1)$ to generate the edited image. We evaluate the performance on the EditEvalv2 benchmark~\citep{huang2024editeval}.
To assess identity preservation, we compute the CLIP image similarity (C-I)~\cite{hessel2021clipscore} between the edited and input images. To evaluate the faithfulness of the edited images, we measure CLIP text alignment (C-T), CLIP Directional Similarity (C-D), Image Reward (IR)~\cite{xu2024imagereward}, and PickScore (PS)~\cite{kirstain2023pick} based on the edited image prompt. 
The reported PS are compared against IP2P. 
As shown in Tab.~\ref{tab:ip2p-results}, our method shows better alignment with the prompt while preserving the identity of the source image. We observe improvements in both IR and PS which have been observed to better align with human preference~\cite{xu2024imagereward, kirstain2023pick} with slight underperformance in CLIP-T (for \texttt{SD1.5} and \texttt{SD2.1}).
Qualitative results are shown in Fig.~\ref{fig:ip2p-selected-qualitatives}. Our method generates faithful, high-fidelity edited images (rows 1 and 2) whereas IP2P creates distorted images (row 3). 

\begin{table}[!h]
    \begin{minipage}{\linewidth}
    \centering
    \scriptsize
        \begin{tabularx}{\linewidth}{>{\centering\arraybackslash}m{0.27\textwidth} | >{\centering\arraybackslash}m{0.09\textwidth} | >{\centering\arraybackslash}m{0.09\textwidth} | >{\centering\arraybackslash}m{0.09\textwidth} | >{\centering\arraybackslash}m{0.09\textwidth} | >
        {\centering\arraybackslash}m{0.08\textwidth}}
        \toprule
        Method & C-I $\uparrow$ & C-T $\uparrow$ & \makecell{C-D $\uparrow$} & \makecell{IR $\uparrow$} & PS $\uparrow$ \\
        \midrule
        IP2P~\cite{brooks2023instructpix2pix} & 0.909 & \underline{0.294} & 0.174 & -0.510 & $-$ \\
        \textbf{Ours} w/ \texttt{SD1.5} & {0.911} & {0.291} & \textbf{0.186} & \underline{-0.460} & {0.514} \\
        \textbf{Ours} w/ \texttt{SD2.1} & \underline{0.913} & 0.290 & {0.184} & {-0.464} & \underline{0.518} \\
        \textbf{Ours} w/ \texttt{PixArt-$\alpha$} & \textbf{0.915} & \textbf{0.297} & \underline{0.185} & \textbf{-0.363} & \textbf{0.532} \\
        \bottomrule
        \end{tabularx}
        \vspace{-0.5\baselineskip}
        \caption{\textbf{Image Editing with InstructPix2Pix (IP2P)~\cite{brooks2023instructpix2pix}}. We normalize text and image similarity scores: C-I, C-T, and C-D. 
        (\textbf{bold} represents the best, and \underline{underline} represents the second best method.)}
    \label{tab:ip2p-results}
    \end{minipage}
    \vspace{-0.5\baselineskip}
\end{table}

\vspace{-0.25\baselineskip}
\section{Conclusion}
\label{sec:conclusion}
\vspace{-0.25\baselineskip}
We presented a novel, training-free approach to improving the generation quality of a CFG-based fine-tuned conditional diffusion model by replacing the low-quality unconditional noise with richer unconditional noise from a separate pretrained base diffusion model. We validated our approach across a range of diffusion models trained for distinct conditional generation, including image variation~\cite{xu2023versatile}, image editing~\cite{brooks2023instructpix2pix}, novel view synthesis~\cite{liu2023zero}, and video generation~\cite{xing2025dynamicrafter}. Notably, we find that the separate pretrained diffusion model can have different weights and architecture from the original base model.

\vspace{-1.0\baselineskip}
\paragraph{Limitations and Discussions.} Although our method is training-free, it involves loading a second model into memory which increases memory cost. Furthermore, we can no longer parallelize computation as done in CFG, resulting in slight inference time overhead. However, the inference speed is only slightly affected as shown in the \textbf{supplementary}.
Our method significantly improves diverse fine-tuned diffusion models, but is less effective for adapter-based fine-tuning methods such as ControlNet~\cite{zhang2023adding} and GLIGEN~\cite{li2023gligen}, which exhibit less degradation in unconditional priors. Identifying unconditional priors for these advanced fine-tuning techniques would be a valuable future direction.

\vspace{-0.5\baselineskip}
\section*{Acknowledgements.} This work was supported by the IITP grants (RS-2024-00399817, RS-2025-25441313, RS-2025-25443318, RS-2025-02653113); and the Industrial Technology Innovation Program (RS-2025-02317326), all funded by the Korean government (MSIT and MOTIE), as well as by the DRB-KAIST SketchTheFuture Research Center.
{
    \small
    \bibliographystyle{ieeenat_fullname}
    \bibliography{main}
}
\clearpage

\maketitlesupplementary

\appendix
\renewcommand{\thesection}{\Alph{section}}

In this supplementary material, we first provide additional evidence for the fine-tuned models' poor unconditional priors by quantitatively showing that the base model has better unconditional generation quality than the fine-tuned models in Sec.~\ref{sec:unconditional-quantitative}. In Sec.~\ref{sec:exp-details}, we include more details about the experimental setups for Zero-1-to-3, Versatile Diffusion, DiT, DynamiCrafter, and InstructPix2Pix. We include more qualitative results in Sec.~\ref{sec:more-qualitatives} and more ablation studies on the CFG scale in Sec.~\ref{app:cfg-scale-ablation}. Finally, we provide details on the inference speed and memory cost of our method in Sec.~\ref{sec:inference-speed}.

\section{Quantitative Evaluation of Unconditional Samples} \label{sec:unconditional-quantitative}
In the main paper, we argued that the poor unconditional priors from the fine-tuned models degrade the quality of the conditional generation. We qualitatively showed in Fig.~\ref{fig:unconditional-samples} \refofpaper{} that the fine-tuned models exhibit poor unconditional generation quality. In this section, we quantitatively show that the base models have better unconditional generation quality than the fine-tuned models. We unconditionally sample 5000 images from each of \texttt{SD1.4}, \texttt{SD2.1}, \texttt{PixArt-$\alpha$}, Zero-1-to-3, Versatile Diffusion, and InstructPix2Pix, and evaluate the image quality using Inception Score (IS)~\cite{salimans2016inceptionscore}. The results are shown in Tab.~\ref{tab:image-unconditional-results}. We observe that the fine-tuned models indeed have quantitatively worse unconditional generation than the base models. Thus, in the main paper, we proposed replacing the poor unconditional noise from the fine-tuned models with the good unconditional noise from the base model which improves the conditional generation quality.

\begin{table}[h]
\begin{minipage}{\linewidth}
\centering
\scriptsize
\begin{tabularx}{\linewidth}{c | Y}
\toprule
Method & IS $\uparrow$ \\
\midrule
\texttt{SD1.4} & \textbf{14.085} \\
\texttt{SD2.1} & 12.640 \\
\texttt{PixArt-$\alpha$} & 9.224 \\
\midrule
Versatile Diffusion & 2.704 \\
Zero-1-to-3 & 9.140 \\
InstructPix2Pix & 5.852 \\
\bottomrule
\end{tabularx}
\vspace{-1.0\baselineskip}
\caption{\textbf{Image Model Unconditional Generation.} We sample using the unconditional noise predictions from each model. The unconditional samples from \texttt{SD1.4}, \texttt{SD2.1}, and \texttt{PixArt-$\alpha$} are higher quality than those of the fine-tuned models. (\textbf{bold} represents the best performance.)}
\label{tab:image-unconditional-results}
\end{minipage}
\end{table}
\section{Experiment Details} \label{sec:exp-details}
For all experiments, we use the DDIM~\citep{songdenoising} sampler. When applying our method to a base model with a different variance schedule, we use the fine-tuned model’s variance schedule as the reference and choose the base model timestep that yields the closest available variance to the fine-tuned model's variance.
\subsection{Zero-1-to-3~\cite{liu2023zero}}
We evaulate our method using the Google Scanned Objects (GSO) dataset~\cite{gso} which consists of over a thousand scanned objects. We render six views for each object at fixed radii and elevation with azimuths uniformly spaced $60^\circ$ apart from each other. The first view is used as the reference image and Zero-1-to-3 is used to generate the remaining five images for evaluation. We use 50 steps of DDIM and a CFG scale of $\gamma = 5.0$.

\subsection{Versatile Diffusion~\cite{xu2023versatile}}
We use the COCO-Captions~\cite{lin2014microsoft} 2014 validation set as the ground truth dataset. We randomly select 30,000 images from the validation set as input conditions to Versatile Diffusion and compute the FID and $\text{FD}_{\text{DINOv2}}$ against the \emph{full} validation set. We use 50 steps of DDIM and a CFG scale of $\gamma = 2.0$.

\subsection{DiT~\cite{peebles2023dit}}
We sample the images using $\gamma=1.5$ and 50 steps of DDIM. The base model used is DiT-XL/2 trained on ImageNet 256$\times$256~\cite{deng2009imagenet}. The fine-tuning is done on each of the datasets using 20,000 steps with batch size 64 and learning rate $0.0001$. To account for the impact of random variation, we compute the FID three times and report the minimum, as done by~\citet{Karras2022edm}. We provide additional details on each of the dataset below.

\paragraph{SUN397~\cite{xiao2010sun}}
SUN397~\cite{xiao2010sun} is a dataset used for testing algorithms for scene recognition consisting of 108,754 images distributed among 397 categories.

\paragraph{Food101~\cite{bossard2014food}}
Food101~\cite{bossard2014food} consists of 101,000 images split among 101 food categories. Each category contains 250 test images and 750 training images.

\paragraph{Caltech101~\cite{griffin2007caltech}}
Caltech101~\cite{griffin2007caltech} contains images of objects belonging to 101 classes, containing 9,145 images in total. Each class contains between 40 and 800 images with a typical edge length of between 200 and 300 pixels.

\subsection{DynamiCrafter~\cite{xing2025dynamicrafter}}
We sample $256 \times 256$ resolution videos using 50 steps of DDIM with a CFG scale of $\gamma_T = 7.5$ and $\gamma_I = 1.5$. Although the original paper uses a CFG scale of $\gamma_T = \gamma_I = 7.5$, we find that their choice of CFG scale results in mostly static images, as shown in their low dynamic degree of 40.57\% in the VBench benchmark~\cite{huang2024vbench}. In contrast, the baseline DynamiCrafter with our choice of CFG scale has a higher dynamic degree of 59.59\%. 

\subsection{InstructPix2Pix~\cite{brooks2023instructpix2pix}} \label{subsec:ip2p-exp-details}
We evaluate the performance of InstructPix2Pix (IP2P) using the EditEvalv2 benchmark~\cite{huang2024editeval} which consists of 150 high quality images with edits from 7 categories.

IP2P uses a dual text-image CFG formulation:
\begin{align}
    \epsilon_\theta(\V{x}_t, c_I, c_T) = &\;{\color{orange} \epsilon_\theta(\vx_t, \emptyset, \emptyset)} \notag \\
    &+ \gamma_I (\epsilon_\theta(\V{x}_t, c_I, \emptyset) - {\color{orange}\epsilon_\theta(\V{x}_t, \emptyset, \emptyset)}) \notag \\
    &+ \gamma_T (\epsilon_\theta(\V{x}_t, c_I, c_T) - \epsilon_\theta(\V{x}_t, c_I, \emptyset)) \label{eq:ip2p-dual-cfg}
\end{align}
For our method, we replace the IP2P \emph{fully} unconditional score ${\color{orange}\epsilon_\theta (x_t, \emptyset, \emptyset)}$ with the unconditional score from \texttt{SD1.5} or \texttt{SD2.1}. We use 100 steps of DDIM with a CFG scale of $\gamma_I = 1.5$ and $\gamma_T = 7.5$. \\
\vspace{-\baselineskip}
\section{Choice of CFG Scale}
\label{app:cfg-scale-ablation}
In this section, we provide an ablation study on the choice of CFG scale $\gamma$ for Zero-1-to-3~\cite{liu2023zero} and Versatile Diffusion~\cite{xu2023versatile}. The results shown in Tab.~\ref{tab:zero123-hyperparameter-results} and~\ref{tab:vd-hyperparameter-results} indicate that our method consistently yields improved results across CFG scales.
\begin{table}[h]
\begin{minipage}{\linewidth}
\centering
\scriptsize
\begin{tabularx}{\linewidth}{c | Y Y Y Y Y Y}
\toprule
$\gamma$ & 3.0 & 4.0 & 5.0 & 6.0 & 7.0 & 8.0 \\
\midrule
Zero-1-to-3~\cite{liu2023zero} & 0.192 & 0.170 & 0.182 & 0.179 & 0.178 & 0.178 \\
\textbf{Ours} w/ \texttt{SD1.4} & \underline{0.170} & \underline{0.165} & \underline{0.163} & \underline{0.163} & \underline{0.161} & \underline{0.161} \\
\textbf{Ours} w/ \texttt{SD2.1} & \textbf{0.165} & \textbf{0.161} & \textbf{0.158} & \textbf{0.159} & \textbf{0.158} & \textbf{0.160} \\
\textbf{Ours} w/ \texttt{PixArt-$\alpha$} & 0.173 & 0.171 & 0.169 & 0.168 & 0.171 & 0.170 \\
\bottomrule
\end{tabularx}
\vspace{-1.0\baselineskip}
\caption{\textbf{Zero-1-to-3~\cite{liu2023zero} (CFG Scales).} We report the LPIPS~\cite{zhang2018perceptual} (lower is better) of applying our method to Zero-1-to-3 using various CFG scales (\textbf{bold} represents the best, and \underline{underline} represents the second best method).}
\label{tab:zero123-hyperparameter-results}
\end{minipage}
\end{table}
\begin{table}[h]
\begin{minipage}{\linewidth}
\centering
\scriptsize
\begin{tabularx}{\linewidth}{c | Y Y Y Y Y}
\toprule
$\gamma$ & 2.5 & 3.0 & 4.0 & 5.0 & 7.5 \\
\midrule
Versatile Diffusion~\cite{xu2023versatile} & 37.96 & 40.19 & 42.07 & 42.33 & 44.80 \\
\textbf{Ours} w/ \texttt{SD1.4} & \textbf{35.67} & \textbf{35.24} & \textbf{35.45} & \textbf{35.60} & \textbf{36.07} \\
\textbf{Ours} w/ \texttt{SD2.1} & 38.29 & \underline{37.44} & \underline{37.83} & \underline{38.44} & \underline{37.71} \\
\textbf{Ours} w/ \texttt{PixArt-$\alpha$} & \underline{37.55} & 39.03 & 39.62 & 40.24 & 40.89 \\
\bottomrule
\end{tabularx}
\vspace{-1.0\baselineskip}
\caption{\textbf{Versatile Diffusion~\cite{xu2023versatile} (CFG Scales).} We report the FID-5k (lower is better) of applying our method to Versatile Diffusion using various CFG scales  (\textbf{bold} represents the best, and \underline{underline} represents the second best method).}
\label{tab:vd-hyperparameter-results}
\end{minipage}
\end{table}
\newpage
\section{Memory and Inference Speed}
\label{sec:inference-speed}
As shown in Tab.~\ref{tab:memory-time-cost}, the inference speed is only slightly affected by our method.
\begin{table}[h]
\begin{minipage}{\linewidth}
\centering
\scriptsize
\begin{tabularx}{\linewidth}{c | Y  Y | Y  Y}
\toprule
\multirow{2}{*}{Method}  & \multicolumn{2}{c|}{Memory (GB)} & \multicolumn{2}{c}{Speed (seconds/sample)} \\
 & Baseline & \textbf{Ours} & Baseline & \textbf{Ours} \\
\midrule
Zero-1-to-3~\citep{liu2023zero} & 4.93 & 10.06 & 2.92 & 3.59 \\
VD & 5.68 & 10.80 & 7.20 & 8.17 \\
DiT & 3.11 & 5.65 & 4.24 & 4.96 \\
IP2P & 5.13 & 10.14 & 19.45 & 21.43 \\
DynamiCrafter & 19.17 & 29.03 & 125.15 & 142.84 \\
\bottomrule
\end{tabularx}
\vspace{-1.3\baselineskip}
\caption{Memory and Inference Speed on an RTX3090 using float32 precision.}
\label{tab:memory-time-cost}
\end{minipage}
\vspace{-1.2\baselineskip}
\end{table}
\section{Additional Qualitative Results} \label{sec:more-qualitatives}
\begin{figure*}
    \centering
    \setlength{\tabcolsep}{0pt}
    \scriptsize
    \begin{tabularx}{\linewidth}{Y | Y | Y Y Y Y}
        \toprule
        \makecell{Input\\Image} & \makecell{Ground\\Truth} & \makecell{Zero-1-to-3\\\textbf{(Baseline)}} & \makecell{w/ \texttt{SD1.4}\\\textbf{(Ours)}} & \makecell{w/ \texttt{SD2.1}\\\textbf{(Ours)}} & \makecell{w/ \texttt{PixArt-$\alpha$}\\\textbf{(Ours)}} \\
        \midrule
        \multicolumn{6}{c}{\includegraphics[width=\linewidth]{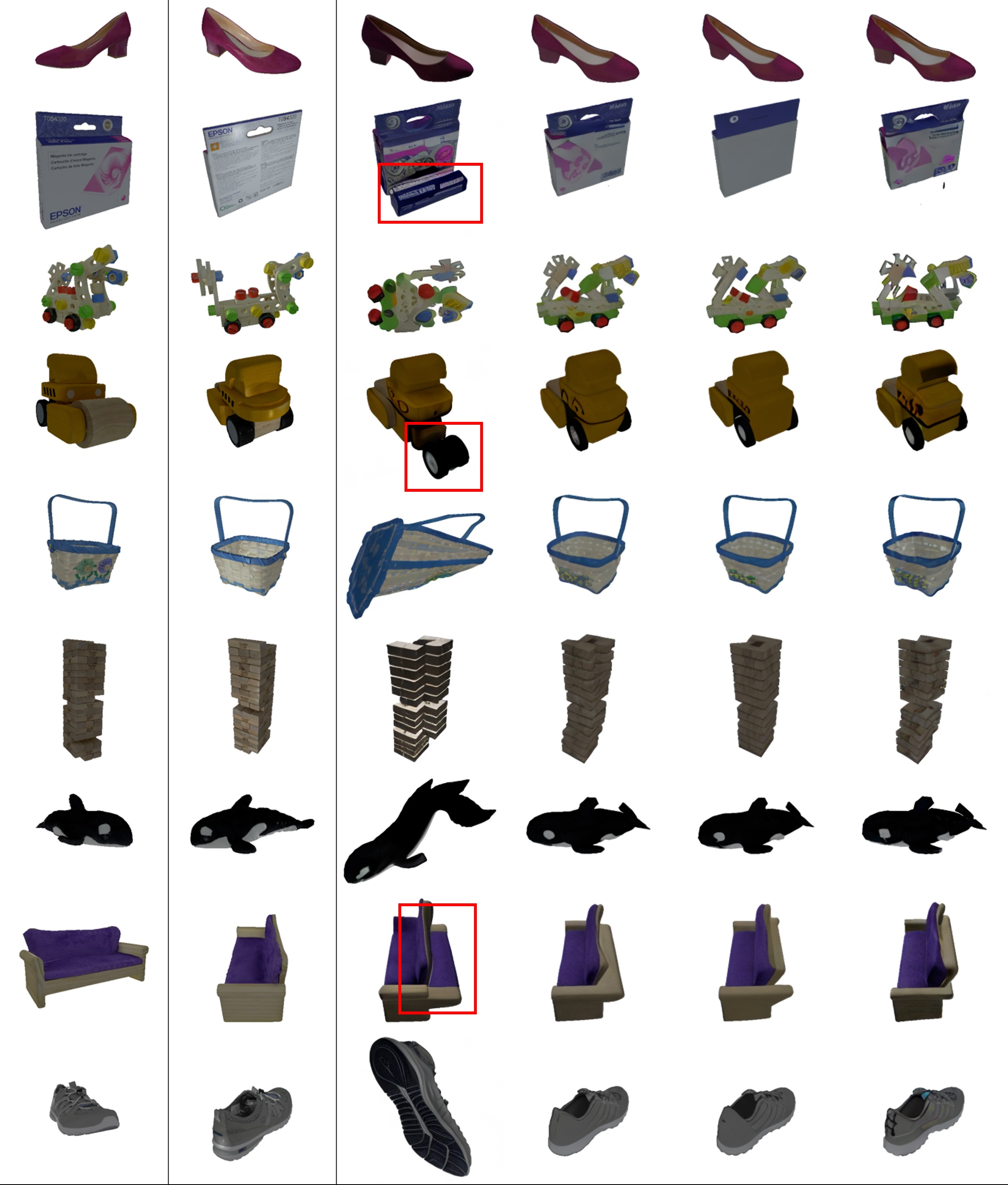}} \\
        \bottomrule
    \end{tabularx}
    \vspace{-\baselineskip}
    \caption{\textbf{Novel View Synthesis with Zero-1-to-3~\cite{liu2023zero}}. Zero-1-to-3 tends to produce views that have inaccurate lighting, coloring, or shape. Combining Zero-1-to-3 with the unconditional noise from \texttt{SD1.4}, \texttt{SD2.1}, or \texttt{PixArt-$\alpha$} corrects these inaccuracies.}
    \label{fig:zero123-qualitatives-supp}
    \vspace{-0.5\baselineskip}
\end{figure*}

\begin{figure*}
    \centering
    \scriptsize
    \setlength{\tabcolsep}{0pt}
    \begin{tabularx}{\textwidth}{Y | Y Y Y Y}
        \toprule
        {\makecell{Input Image}} & \makecell{VD\\\textbf{(Baseline)}} & \makecell{w/ \texttt{SD1.4}\\\textbf{(Ours)}} & \makecell{w/ \texttt{SD2.1}\\\textbf{(Ours)}} & \makecell{w/ \texttt{PixArt-$\alpha$}\\\textbf{(Ours)}} \\
        \midrule
        \multicolumn{5}{c}{\includegraphics[width=\textwidth]{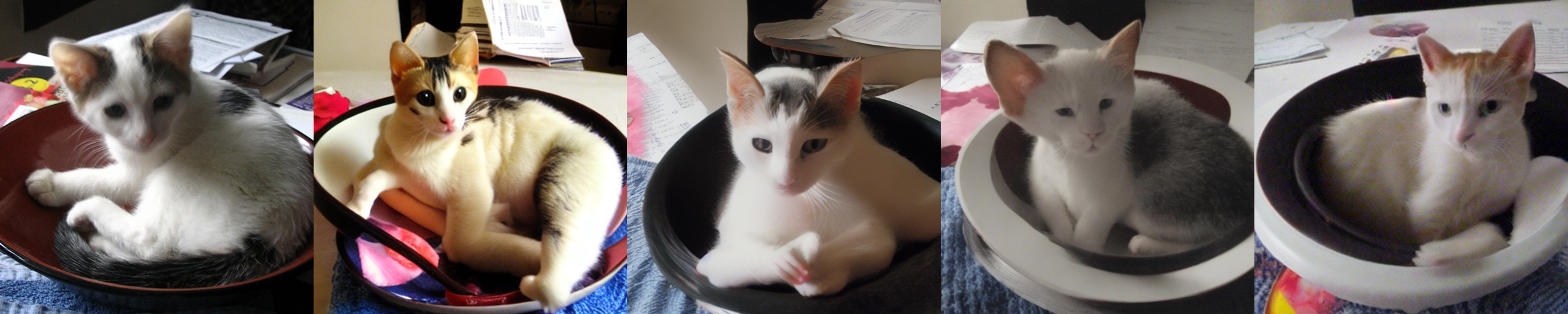}} \\
        \multicolumn{5}{c}{\includegraphics[width=\textwidth]{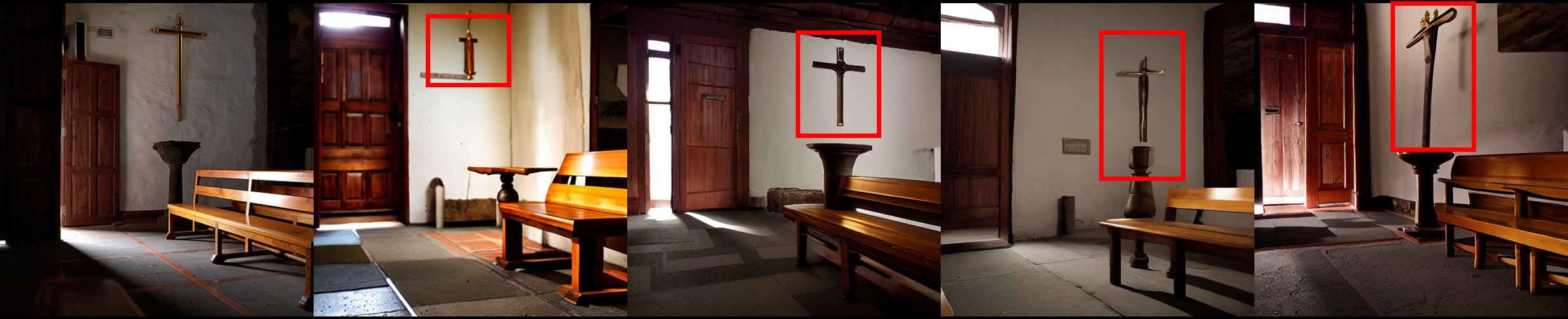}} \\
        \multicolumn{5}{c}{\includegraphics[width=\textwidth]{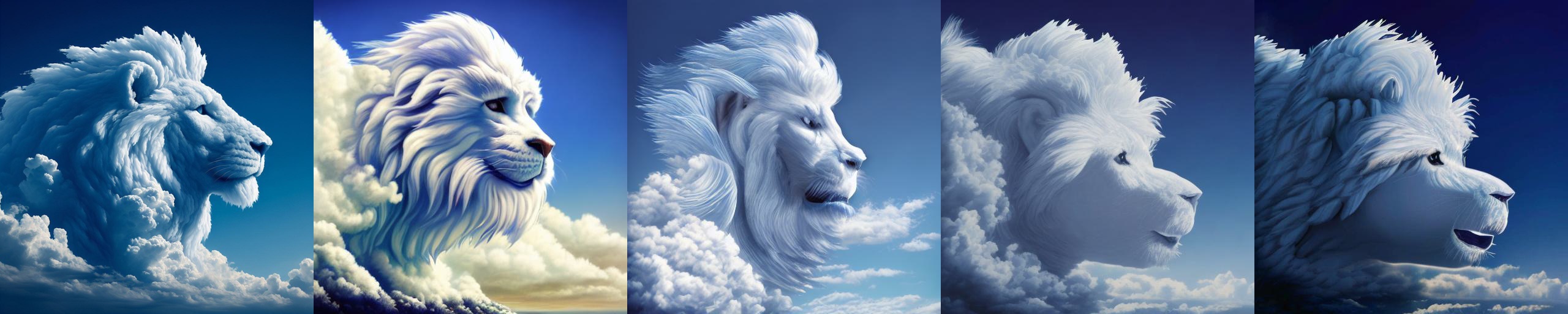}} \\
        \multicolumn{5}{c}{\includegraphics[width=\textwidth]{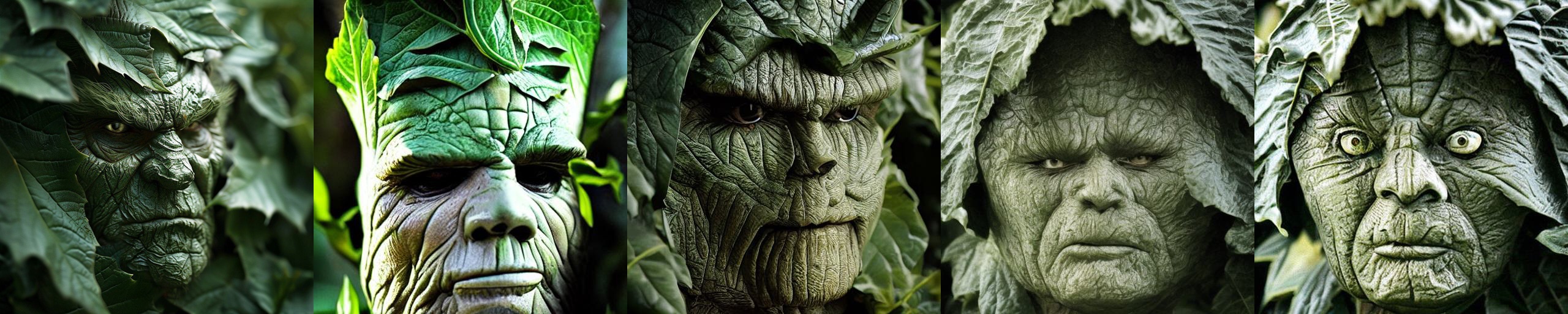}} \\
        \multicolumn{5}{c}{\includegraphics[width=\textwidth]{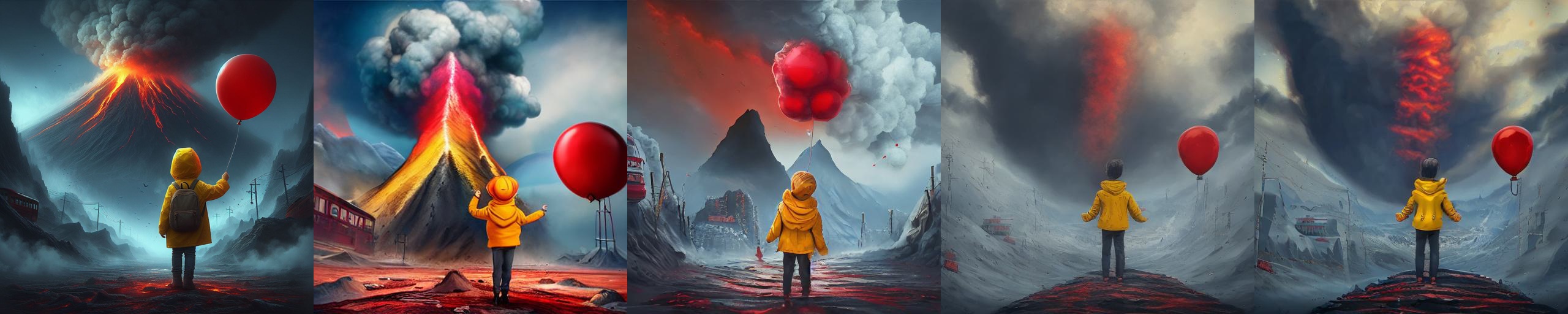}} \\
        \multicolumn{5}{c}{\includegraphics[width=\textwidth]{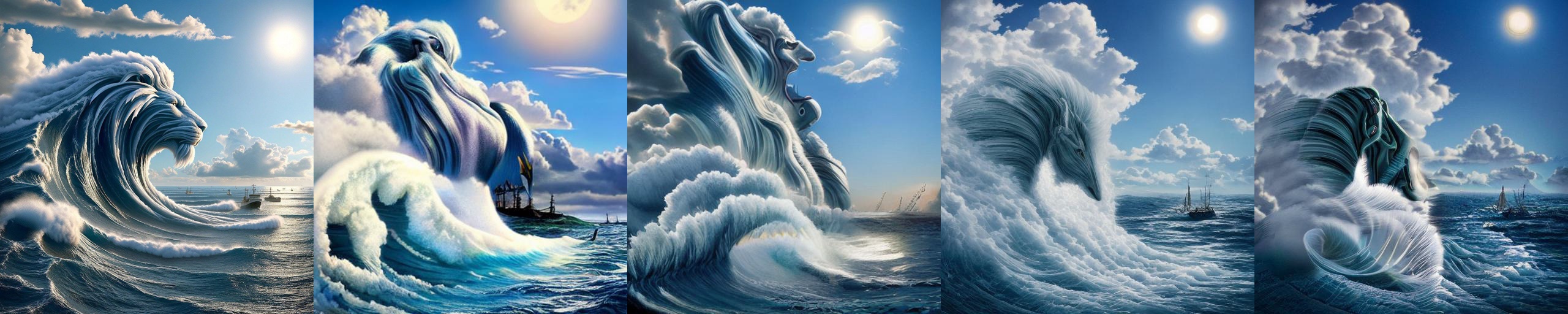}} \\
        \bottomrule
    \end{tabularx}
    \vspace{-\baselineskip}
    \caption{\textbf{Image Variations with Versatile Diffusion~\cite{xu2023versatile}.} Images generated from Versatile Diffusion tend to be oversaturated and distorted. Combining Versatile Diffusion with the unconditional noise predictions from \texttt{SD1.4}, \texttt{SD2.1}, or \texttt{PixArt-$\alpha$} corrects these artifacts.}
    \label{fig:vdiv-supp-qualitatives}
    \vspace{-1.0\baselineskip}
\end{figure*}

\begin{figure*}
    \centering
    \includegraphics[width=\linewidth]{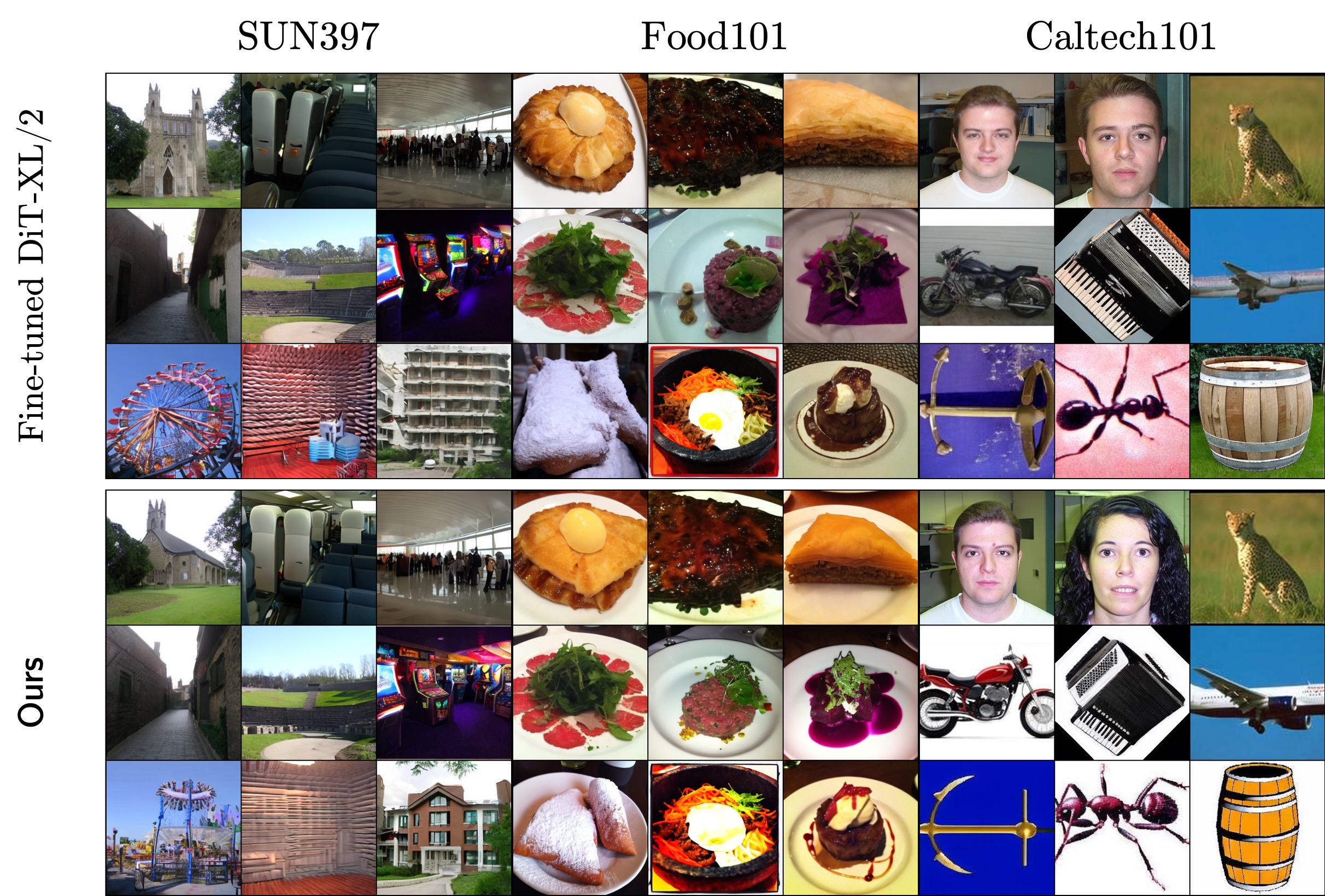}
    \caption{\textbf{Class-conditional generation with DiT~\cite{peebles2023dit}}. Class-conditional generation using DiT fine-tuned on SUN397~\cite{xiao2010sun}, Food101~\cite{bossard2014food}, and Caltech101~\cite{griffin2007caltech}.}
    \label{fig:dit-supp-qualitatives}
\end{figure*}

\begin{figure*}[h]
    \centering
    \vspace{-\baselineskip}
    \setlength{\tabcolsep}{0pt}
    \begin{tabularx}{\linewidth}{>{\centering\arraybackslash}m{0.1835\textwidth} | Y Y Y Y}
        \toprule
        Input & \multicolumn{4}{c}{\textbf{Generated Frames}} \\
        \midrule 
        \multicolumn{5}{c}{\includegraphics[width=\linewidth]{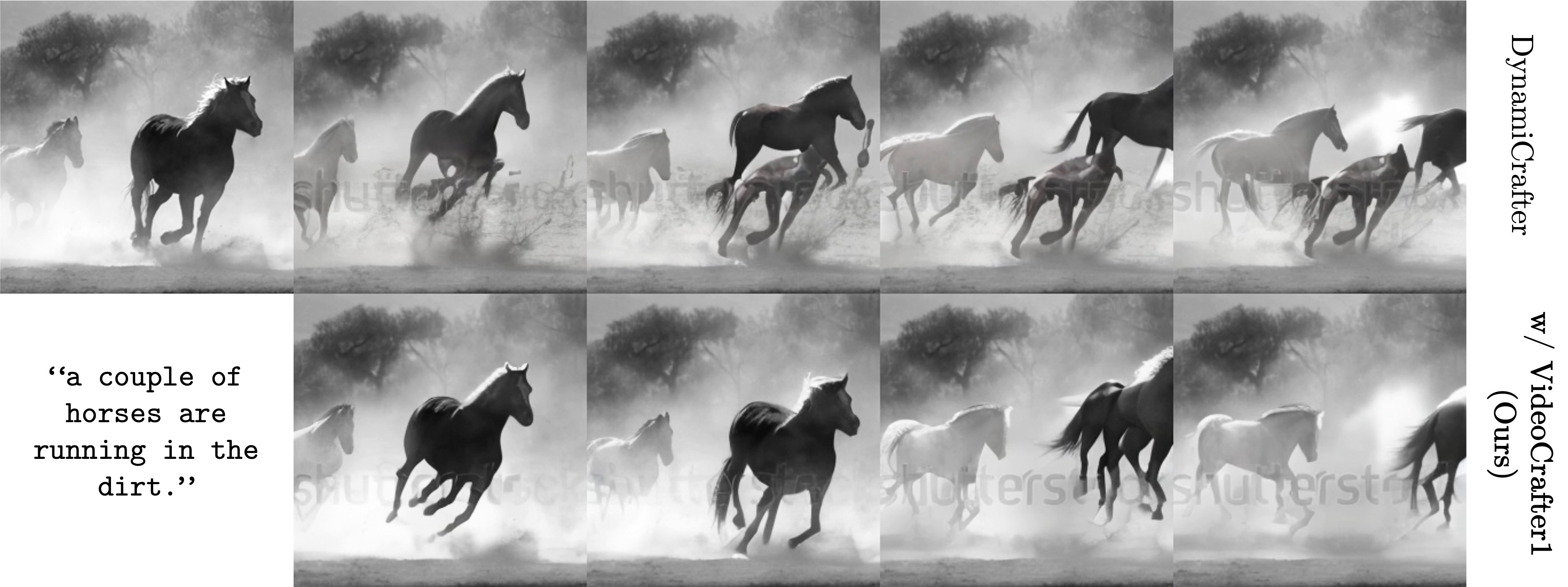}} \\
        \multicolumn{5}{c}{\includegraphics[width=\linewidth]{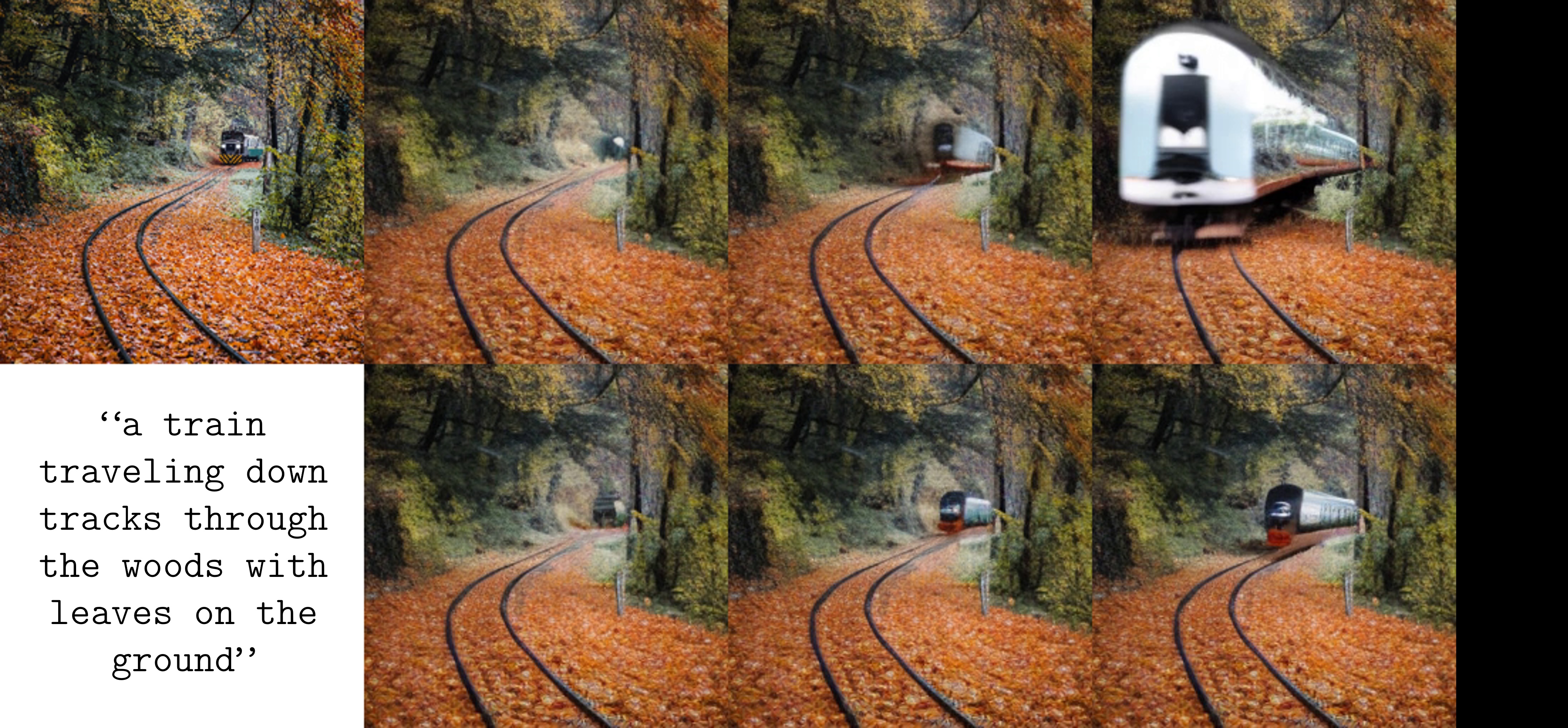}} \\
        \multicolumn{5}{c}{\includegraphics[width=\linewidth]{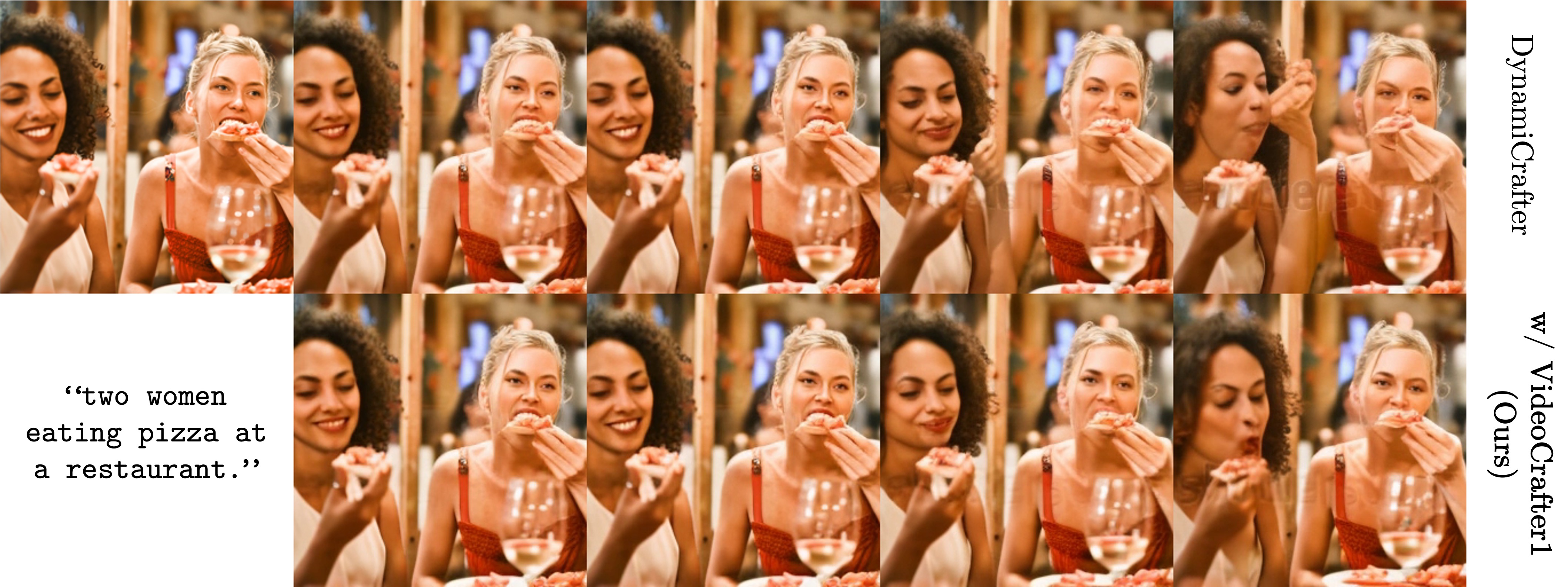}} \\
        \bottomrule
    \end{tabularx}
    \caption{\textbf{Image-to-Video Generation with DynamiCrafter~\cite{xing2025dynamicrafter}.} Our method is more temporally consistent (number of horses in the first video and train color in the second video) and less distorted (hand and face in the last video).}
    \label{fig:dynamicrafter-supp-qualitatives}
\end{figure*}

\begin{figure*}
    \centering
    \setlength{\tabcolsep}{0pt}
    \scriptsize
    \vspace{-0.25\baselineskip}
    \begin{tabularx}{\linewidth}{Y | Y Y Y Y}
        \toprule
        {\makecell{Input \\ Image}} & {\makecell{IP2P\\\textbf{(Baseline)}}} & {\makecell{w/ \texttt{SD1.5}\\\textbf{(Ours)}}} &
        {\makecell{w/ \texttt{SD2.1}\\\textbf{(Ours)}}} &  {\makecell{w/ \texttt{PixArt-$\alpha$}\\\textbf{(Ours)}}} \\
        \midrule
        \multicolumn{5}{c}{\scriptsize\texttt{``\input{figures/ip2p/piebench/03.txt}''}} \\
        \multicolumn{5}{c}{\includegraphics[width=\linewidth]{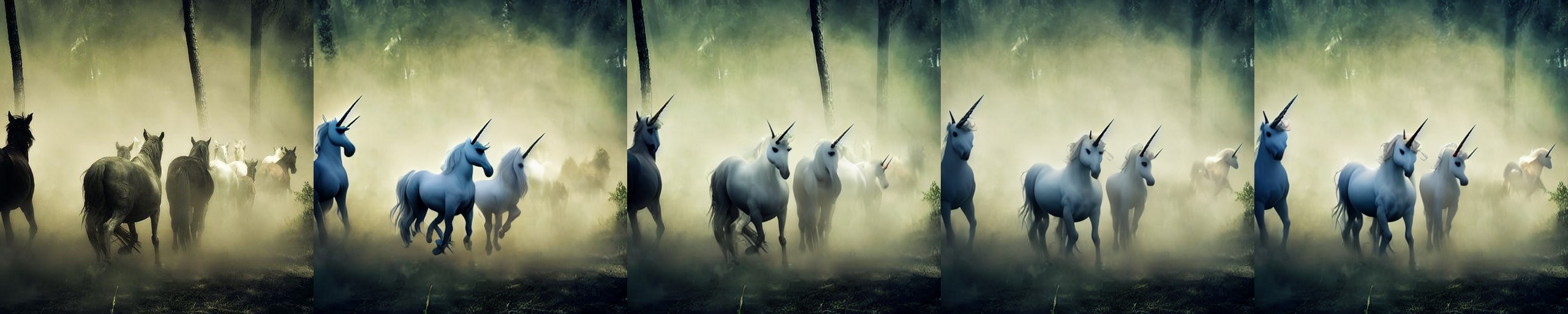}} \\
        \multicolumn{5}{c}{\scriptsize\texttt{``\input{figures/ip2p/piebench/07.txt}''}} \\
        \multicolumn{5}{c}{\includegraphics[width=\linewidth]{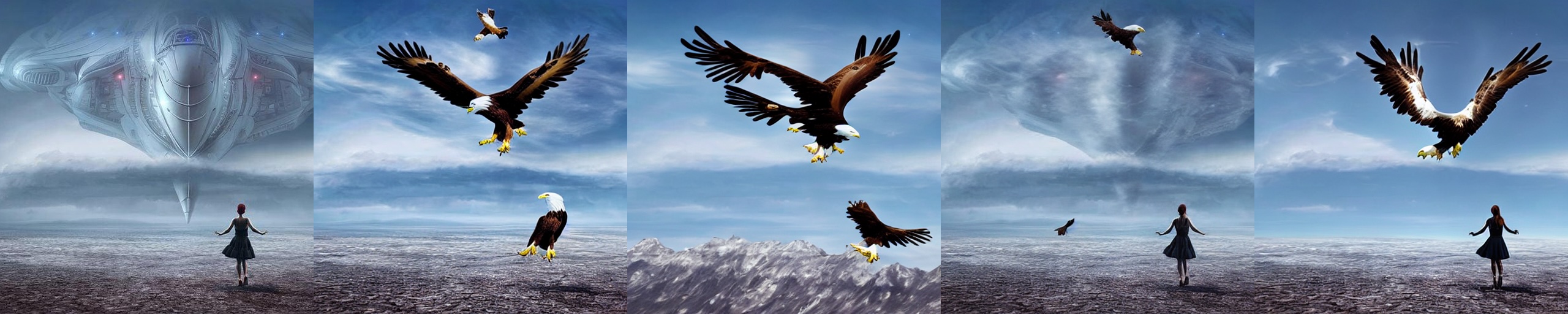}} \\
        \multicolumn{5}{c}{\scriptsize\texttt{``\input{figures/ip2p/piebench/04.txt}''}} \\
        \multicolumn{5}{c}{\includegraphics[width=\linewidth]{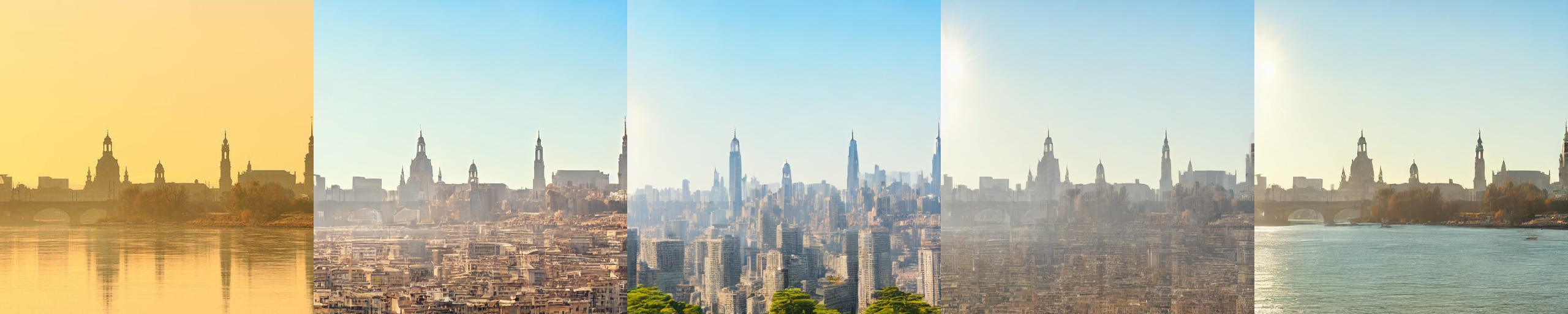}} \\
        \multicolumn{5}{c}{\scriptsize\texttt{``\input{figures/ip2p/piebench/06.txt}''}} \\
        \multicolumn{5}{c}{\includegraphics[width=\linewidth]{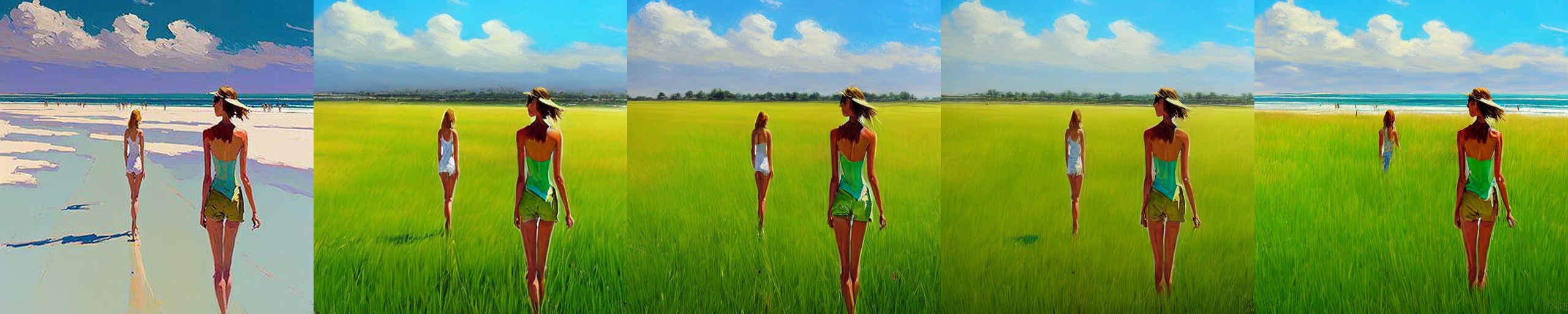}} \\
        \multicolumn{5}{c}{\scriptsize\texttt{``\input{figures/ip2p/piebench/12.txt}''}} \\
        \multicolumn{5}{c}{\includegraphics[width=\linewidth]{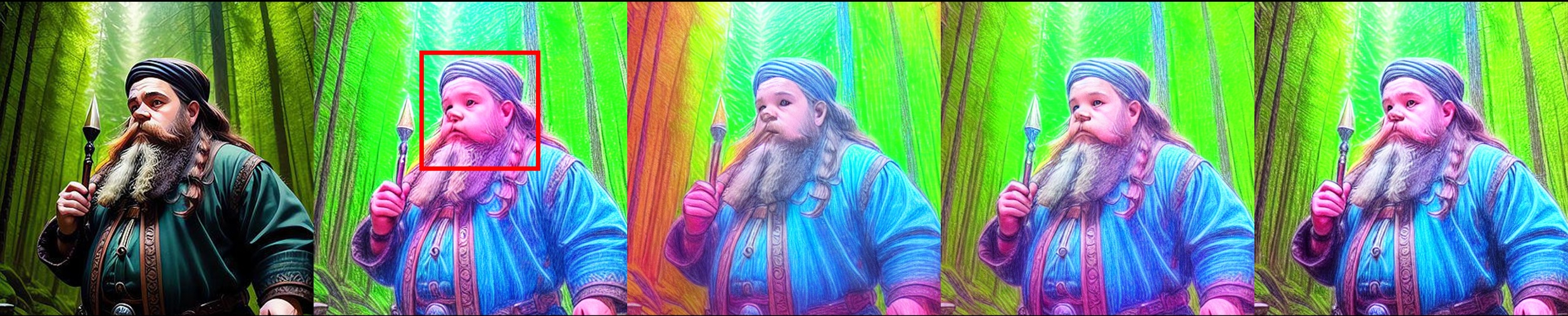}} \\
        \bottomrule
    \end{tabularx}
    \vspace{-\baselineskip}
    \caption{\textbf{Image Editing with InstructPix2Pix (IP2P)~\cite{brooks2023instructpix2pix}.} InstructPix2Pix tends to produce distorted edits. Replacing the IP2P fully unconditional noise with the unconditional noise from \texttt{SD1.5}, \texttt{SD2.1}, or \texttt{PixArt-$\alpha$} corrects these distortions and improves image quality.}
    \label{fig:ip2p-supp-qualitatives}
\end{figure*}

We provide additional qualitative results for Zero-1-to-3 (Fig.~\ref{fig:zero123-qualitatives-supp}), Versatile Diffusion (Fig.~\ref{fig:vdiv-supp-qualitatives}), DiT (Fig.~\ref{fig:dit-supp-qualitatives}), DynamiCrafter (Fig.~\ref{fig:dynamicrafter-supp-qualitatives}), and InstructPix2Pix (Fig.~\ref{fig:ip2p-supp-qualitatives}).
\clearpage
\clearpage
\clearpage
\clearpage
\clearpage

\end{document}